\documentclass[10pt,journal,compsoc]{IEEEtran}

\ifCLASSOPTIONcompsoc
  \usepackage[nocompress]{cite}
\else
  \usepackage{cite}
\fi
\ifCLASSINFOpdf
\else
\fi
\usepackage{booktabs}
\usepackage{subcaption}
\usepackage{multirow}
\usepackage{mathtools}
\usepackage{hanging}
\usepackage{tablefootnote}
\usepackage{epsfig}
\usepackage{graphicx}
\usepackage{bmpsize}
\usepackage{amsmath}
\usepackage{amssymb}
\usepackage{url}
\usepackage{bbding}
\usepackage{makecell}
\usepackage[dvipsnames]{xcolor}
\usepackage{color}
\usepackage{colortbl}

\begin{document}
%
\title{Accurate and Efficient Stereo Matching via Attention Concatenation Volume}
%
%
%
%

\author{Gangwei Xu, Yun Wang, Junda Cheng, Jinhui Tang,~\IEEEmembership{Member,~IEEE}, Xin Yang,~\IEEEmembership{Member,~IEEE}
\IEEEcompsocitemizethanks{\IEEEcompsocthanksitem G. Xu, Y. Wang, J. Cheng and X. Yang are with the Department
of Electronic Information and Communications,
Huazhong University of Science and Technology, Wuhan 430074, China (e-mail:\{gwxu, wangyun, cjd, xinyang2014\}@hust.edu.cn)
\IEEEcompsocthanksitem J. Tang is with Nanjing University of Science and Technology (e-mail: jinhuitang@njust.edu.cn).}
\thanks{Corresponding author: Xin Yang.}}

\IEEEtitleabstractindextext{%
\begin{abstract}
Stereo matching is a fundamental building block for many vision and robotics applications. An informative and concise cost volume representation is vital for stereo matching of high accuracy and efficiency. In this paper, we present a novel cost volume construction method, named attention concatenation volume (ACV), which generates attention weights from correlation clues to suppress redundant information and enhance matching-related information in the concatenation volume. The ACV can be seamlessly embedded into most stereo matching networks, the resulting networks can use a more lightweight aggregation network and meanwhile achieve higher accuracy. We further design a fast version of ACV to enable real-time performance, named Fast-ACV,  which generates high likelihood disparity hypotheses and the corresponding attention weights from low-resolution correlation clues to significantly reduce computational and memory cost and meanwhile maintain a satisfactory accuracy. The core ideas of our Fast-ACV comprise Volume Attention Propagation (VAP) and Fine-to-Important (F2I) strategy. The VAP can automatically select accurate correlation values from an interpolated correlation volume and propagate these accurate values to the surrounding pixels with ambiguous correlation clues, and the F2I can generate a set of disparity hypotheses with high likelihood and the corresponding attention weights to significantly suppress impossible disparities in the concatenation volume and in turn reduce time and memory cost. Furthermore, we design a highly accurate network ACVNet and a real-time network Fast-ACVNet based on our ACV and Fast-ACV respectively, which achieve state-of-the-art performance on several benchmarks (i.e., our ACVNet ranks the $2^{nd}$ on KITTI 2015 and Scene Flow, and the $3^{rd}$ on KITTI 2012 and ETH3D among all the published methods; our Fast-ACVNet outperforms almost all state-of-the-art real-time methods on Scene Flow, KITTI 2012 and 2015 and meanwhile has better generalization ability). The source code is available at \textcolor{magenta}{https://github.com/gangweiX/ACVNet} and \textcolor{magenta}{https://github.com/gangweiX/Fast-ACVNet}. 
\end{abstract}

}

\maketitle

\IEEEdisplaynontitleabstractindextext

%
\IEEEpeerreviewmaketitle

\IEEEraisesectionheading{\section{Introduction}\label{sec:introduction}}

%
%
%
%
\IEEEPARstart{S}{tereo} matching, which estimates depth (or disparity) from a pair of rectified stereo images~\cite{hirschmuller2007stereo, scharstein2002taxonomy}, is a fundamental task for many robotics and computational photography applications, such as 3D reconstruction, robot navigation and autonomous driving. Despite a plethora of research works in the literature, stereo matching remains challenging due to difficulties in tackling repetitive structures, texture-less/transparent objects and occlusions. Meanwhile, how to concurrently achieve high inference accuracy and efficiency is critical for practical applications yet remains challenging.

Recently, convolutional neural networks have exhibited great potential in this field. State-of-the-art CNN stereo models typically consist of four steps, i.e. feature extraction, cost volume construction, cost aggregation and disparity regression. The cost volume which provides initial similarity measures for left image pixels and possible corresponding right image pixels is a crucial step of stereo matching. An informative and concise cost volume representation from this step is vital for the final accuracy and computational complexity. Learning-based methods explore different cost volume representations. DispNetC~\cite{dispNetC2016large} computes a single-channel full correlation volume between the left and right feature maps along every disparity level.  Such full correlation volume provides an efficient way for measuring similarities, but it loses much content information. GC-Net~\cite{kendall2017end} constructs a 4D concatenation volume by concatenating left and right feature maps along all disparity levels to provide abundant content information. However, the concatenation volume does not provide explicit similarity measurements and thus requires extensive 3D convolutions for cost aggregation to learn similarity measurements from scratch. To tackle the above drawbacks, GwcNet~\cite{guo2019group} concatenates the group-wise correlation volume with a compact concatenation volume to encode both matching and content information in the final 4D cost volume. However, the data distribution and characteristics of a correlation volume and a concatenation volume are quite different, i.e. the former represents the similarity measurement obtained through the dot product, and the latter is the concatenation of the unary features. Simply concatenating the two volumes and regularizing them via 3D convolutions can hardly exert the advantages of the two volumes to the full. As a result, GwcNet~\cite{guo2019group} still requires extensive 3D convolutions for cost aggregation. To further reduce the memory and computational complexity, several methods ~\cite{shen2021cfnet, gu2020cascade,cheng2020deep} employ the cascade cost volume which builds a cost volume pyramid in a coarse-to-fine manner to progressively narrow down the target disparity range. However, these cascaded methods need to re-construct and re-aggregate a cost volume for each stage without reusing prior information in the probability volume of the previous stage, yielding a low utilization efficiency. In addition, these cascaded methods could suffer from irreversible cumulative errors as they directly discard disparities that are beyond the prediction range in the previous stages.

To balance efficiency and accuracy, several recent studies attempt to construct and aggregate only a low-resolution (i.e., 1/8) 4D cost volume and restore a full resolution disparity map via up-sampling. For instance, StereoNet~\cite{stereonet2018} constructs a 4D cost volume based on the differences between the left feature and right feature maps at only 1/8 resolution. Then 2D disparity maps are regressed and then up-sampled via bilinear interpolation from the 1/8 resolution cost volume. Similarly, BGNet~\cite{xu2021bilateral} constructs a low-resolution 4D  group-wise correlation volume and designs a parameter-free slicing layer based on a bilateral grid to obtain an edge-preserving high-resolution cost volume from the low-resolution cost volume. DeepPruner~\cite{deeppruner2019} develops a differentiable PatchMatch~\cite{patchmatch2014} module to efficiently construct a sparse representation of a low-resolution concatenation volume. The search space of each pixel is pruned by the predicted minimum and maximum disparities. Unfortunately, these efficiency-oriented methods typically degrade the accuracy greatly compared with the best-performing algorithms.

This work aims to explore a more efficient and effective form of cost volume, which can achieve state-of-the-art accuracy with high efficiency. We build our model based on two key observations: First, the correlation volume which measures feature similarities between left and right images can quickly provide rough geometric structure information. Second, the concatenation volume contains rich content and fine structure but redundant information. This suggests that utilizing the correlation volume which encodes geometric structure information can facilitate a concatenation volume to significantly suppress its redundant information and meanwhile maintain sufficient information for estimating the correct disparity.

\begin{figure*}[t]
    \centering
    \includegraphics[width=0.9\linewidth]{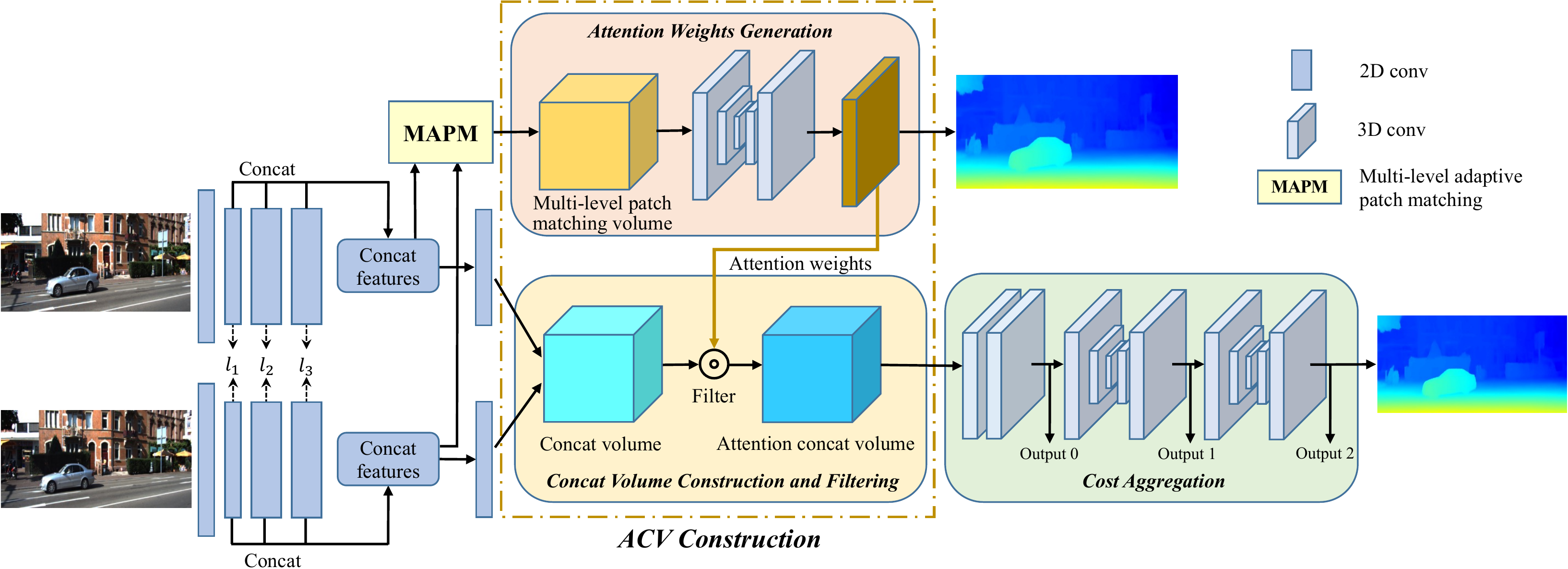}
    \caption{The structure of our proposed ACVNet. The construction process of ACV consists of three steps: Attention Weights Generation, Initial Concatenation Volume Construction and Attention Filtering. We obtain feature maps at three different levels $l_1$, $l_2$ and $l_3$ from the feature extraction module, and the number of channels for $l_1$, $l_2$ and $l_3$ is 64, 128 and 128 respectively. $l_1$, $l_2$ and $l_3$ are concatenated to form 320-channel concat features for the generation of attention weights. Then two convolutions are applied to compress the 320-channel concat features to 32-channel features for construction of the initial concatenation volume.
    }
    \label{fig:acvnet}
    \vspace{-10pt}
\end{figure*}

With these intuitions in mind, we propose an Attention Concatenation Volume (ACV) which exploits a lightweight correlation volume to generate attention weights to filter a concatenation volume (see Fig. \ref{fig:acvnet}). The ACV can achieve high accuracy and meanwhile significantly alleviate the burden of cost aggregation. Experimental results show that after replacing the combined volume of GwcNet with our ACV, only four 3D convolutions for cost aggregation can achieve better accuracy than GwcNet which employs twenty-eight 3D convolutions for cost aggregation. Our ACV is a general cost volume representation that can be seamlessly integrated into various 3D CNN stereo models for performance improvement. Results show that after applying our method, PSMNet~\cite{chang2018pyramid} and GwcNet~\cite{guo2019group} can respectively achieve 28\% and 39\% accuracy improvements.

We further design a faster version of ACV to enable real-time performance (i.e., inference time of stereo matching is less than 50ms), called Fast-ACV (see Fig. \ref{fig:fast_acvnet}). The key differences between Fast-ACV and ACV are three folds. First, we propose a novel Volume Attention Propagation (VAP) block which effectively restores a high-quality correlation volume of high resolution from the low-resolution volume. This VAP block is seamlessly integrated into the attention weights generation module. By doing so, we only need to construct a low-resolution correlation volume, leading to significant reduction in the time required for correlation volume construction and aggregation. Compared to the cost volume-based linear interpolation (which relies on pixel distance), our VAP can adaptively perform accurate interpolation based on both content information and confidence. This adaptive approach minimizes the ambiguity inherent in linear interpolation, consequently enhancing the accuracy of the cost volume. The VAP is a novel cost volume interpolation method that delivers noteworthy performance improvements with negligible cost. These findings are poised to stimulate future research in the field of interpolation methods.
Specifically, our VAP automatically identifies a set of pixels with reliable correlation values in an interpolated correlation volume of high resolution and propagates such information to their neighbors to progressively revise errors and reduce ambiguities in the high-resolution correlation volume. Pixels that have accurate regressed disparities and sharply-distributed disparity probabilities (i.e. high confidence of regressed disparity) are considered to be reliable and can be used to guide the revision of correlation values for the neighboring pixels. To this end, we design an overall measure to evaluate the disparity estimation accuracy and confidence for each pixel in the interpolated correlation volume. In addition, we adopt a cross shape sampling pattern to enable a highly effective and efficient propagation path. Second, we propose a Fine-to-Important (F2I) sampling strategy which generates a set of disparity hypotheses with high likelihood and the corresponding attention weights to significantly suppress impossible disparities in the concatenation volume and in turn reduce time and memory cost with little accuracy degradation. Third, we utilize a lightweight backbone for feature extraction and a lightweight aggregation network as most efficiency-oriented stereo models for real-time performance.

Based on the advantages of the proposed ACV and Fast-ACV, we design an accurate stereo matching network ACVNet and its real-time version Fast-ACVNet. At the time of writing, our ACVNet ranks the $2^{nd}$ on KITTI 2015~\cite{kitti2015} and Scene Flow~\cite{dispNetC2016large}, and the $3^{rd}$ on KITTI 2012~\cite{kitti2012} and ETH3D~\cite{schops2017multi} among all the published methods. It is noteworthy that our ACVNet is the only method that ranks top 3 concurrently on all four datasets above, demonstrating its good generalization ability to various scenes. Regarding the inference speed, our ACVNet is the fastest among the top 10 methods in the KITTI benchmarks.  Meanwhile, our Fast-ACVNet outperforms almost all state-of-the-art real-time methods on Scene Flow~\cite{dispNetC2016large}, KITTI 2012~\cite{kitti2012} and 2015~\cite{kitti2015}.

In summary, our main contributions are:

\begin{itemize}
     \item We propose an Attention Concatenation Volume (ACV) and its real-time version Fast-ACV, which adopt a unified framework to construct highly informative and concise cost volume representation and can be seamlessly integrated into most existing stereo models to significantly reduce computational cost and improve accuracy.
    \item We propose a Volume Attention Propagation module and a Fine-to-Important sampling strategy, which are key success factors of Fast-ACV. 
     \item We propose a highly accurate stereo matching model, ACVNet, and a real-time stereo matching model, Fast-ACVNet, which achieve state-of-the-art performance on several popular benchmarks.
     \item We release the source codes of ACVNet and Fast-ACVNet.
\end{itemize}

\section{Related Work}
\subsection{Cost Volume based Deep Stereo Matching}
Recently, CNN-based stereo models~\cite{chang2018pyramid, guo2019group, liang2019stereo, cspn, wang2020parallax, wu2019semantic, zhang2020adaptive, xu2023iterative} have achieved impressive performance on almost all the standard benchmarks. Most of them devote to improving the accuracy and efficiency of cost volume construction and cost aggregation, which are the two key steps of stereo matching.

\vspace{2mm}
\noindent\textbf{Cost Volume Construction.} Existing cost volume representation can be roughly categorized into three types: correlation volume, concatenation volume and combined volume by concatenating the two volumes. DispNetC~\cite{dispNetC2016large} utilizes a correlation layer to directly measure the similarities of left and right image features to form a single-channel cost volume for each disparity level. Then, 2D convolutions are applied to aggregate contextual information. Such full correlation volume demands low memory and computational complexity, yet the encoded information is too limited (i.e. too much content information is lost in the channel dimension) to achieve a satisfactory accuracy. GC-Net~\cite{kendall2017end} uses the concatenation volume, which concatenates the left and right CNN features to form a 4D cost volume for all disparities. Such 4D concatenation volume preserves abundant content information from all feature channels and thus outperforms the correlation volume in terms of accuracy. However, as the concatenation volume does not explicitly encode similarity measures, it requires a deep stack of 3D convolutions to aggregate costs of all disparities from scratch. To overcome the above drawbacks, GwcNet~\cite{guo2019group} proposes the group-wise correlation volume and concatenates it with a compact concatenation volume to form a combined volume, which aims to combine the advantages of two volumes. However, directly concatenating two types of volumes without considering their respective characteristics yields an inefficient use of the complementary strengths in the two volumes. As a result, deep stacking 3D convolutions in the hourglass architecture are still demanded for cost aggregation in GwcNet~\cite{guo2019group}. Following the 4D combined cost volume, cascaded approaches ~\cite{gu2020cascade, shen2021cfnet} build a 4D cost volume pyramid in a coarse-to-fine manner to progressively narrow down the target disparity range and refine the disparity map. However, these cascaded approaches need to reconstruct and aggregate the cost volume for every stage which also incur abundant 3D convolutions. In addition, they usually do not use the prior information in the probability volume of the previous stage, yielding a low utilization efficiency. Moreover, such coarse-to-fine strategy inevitably involves irreversible accumulated errors, i.e., refining only the peak disparity regressed from the previous stage may miss the true disparity when the disparity distribution contains multiple peaks.

\vspace{2mm}
\noindent\textbf{Cost Aggregation.} The goal of this step is to 
aggregate or regularize the cost volume to derive accurate similarity measures. Many existing methods~\cite{chang2018pyramid,guo2019group, mvsnet,cheng2023coatrsnet} exploit deep 3D CNNs to aggregate or regularize the cost volume. However, this step comes with substantial memory demands. To reduce memory cost, R-MVSNet~\cite{rmvsnet} regularizes the cost volume in a sequential manner using the convolutional gated recurrent unit (GRU) rather than 3D CNNs. Nevertheless, this solution comes at the cost of an increase in run-time. To reduce the complexity, AANet~\cite{xu2020aanet} proposes an intra-scale and cross-scale cost aggregation algorithm to replace the conventional 3D convolutions which can achieve very fast inference speed with a sacrifice of nontrivial accuracy degradation. GANet~\cite{zhang2019ga} also tries to replace 3D convolutions with two guided aggregation layers, which achieves a higher accuracy using spatially dependent 3D aggregation at the cost of a higher aggregation time due to the two guided aggregation layers. 

Cost volume construction and aggregation are two tightly coupled modules that jointly determine the accuracy and efficiency of a stereo matching network. In this work, we propose a highly efficient yet informative cost volume representation, named attention concatenation volume, by using the similarity information encoded in the correlation volume to regularize the concatenation volume so that only a lightweight aggregation network is demanded to achieve an overall high efficiency and accuracy.

\subsection{Real-time Stereo Matching}
Several recent studies~\cite{stereonet2018, deeppruner2019, xu2021bilateral,yao2021decomposition} focus on lightweight stereo networks based on 4D cost volumes to achieve real-time performance and meanwhile maintain satisfactory accuracy. These methods typically construct and aggregate a 4D cost volume at low resolution to significantly reduce computational cost. To compensate information loss at low resolution, StereoNet ~\cite{stereonet2018} proposes an edge-preserving refinement network, which utilizes the left images as a guidance to recover high-frequency details. DeepPruner~\cite{deeppruner2019} adopts the idea of PatchMatch~\cite{patchmatch2014} to first build a sparse representation of the cost volume, and then prune the search space based on the predicted minimum and maximum disparities. The predicted disparities are further refined under the guidance of low-level image feature maps. BGNet~\cite{xu2021bilateral} proposes an up-sampling module based on the learned bilateral grid to restore a 4D cost volume of high resolution from a low-resolution cost volume. HITNet~\cite{tankovich2021hitnet} represents image tiles as planar patches and integrates image warping, spatial propagation, and a fast high-resolution initialization step into the network, to reduce computational cost.  However, HITNet~\cite{tankovich2021hitnet} requires extra propagation loss, slant loss and confidence loss for training, which could lead to a poor generalization ability in unseen scenes with different characteristics from the training data. In comparison, our Fast-ACV has better generalization ability and can be applied to many 4D cost volume based stereo models as plugins, which are more convenient. 

Motivated by existing real-time methods, our real-time version of ACV (i.e., Fast-ACV) employs a novel propagation strategy (via the VAP module) to efficiently obtain interpolated correlation values which can better tolerate noises in the original low-resolution correlation volume and meanwhile effectively recover thin structures and sharp boundaries. Meanwhile, different from existing real-time methods which directly regress disparities from the interpolated correlation volume, we use the interpolated correlation values to generate disparity hypotheses with high likelihood and the corresponding attention weights to regularize the concatenation volume. Also, our Fast-ACV is different from the cascaded approaches which narrow down the disparity search space via a coarse-to-fine strategy, our method preserves all disparity hypotheses with high likelihood and adjusts the attention weights of hypotheses (i.e., a fine-to-important strategy) to avoid irreversible cumulative errors.

\section{Method}
In this section, we first describe the basic design of our ACV (Sec. \ref{sec:acv}) and Fast-ACV (Sec. \ref{sec:fast_acv}). Then in Secs. \ref{sec:acvnet_architecture} and \ref{sec:fast_acvnet_architecture} we present details of network architectures of ACVNet and Fast-ACVNet respectively. Finally, in Sec. \ref{sec:loss}, we explain the loss functions used to train our ACVNet and Fast-ACVNet. 

\begin{figure*}[t]
    \centering
    \includegraphics[width=0.9\linewidth]{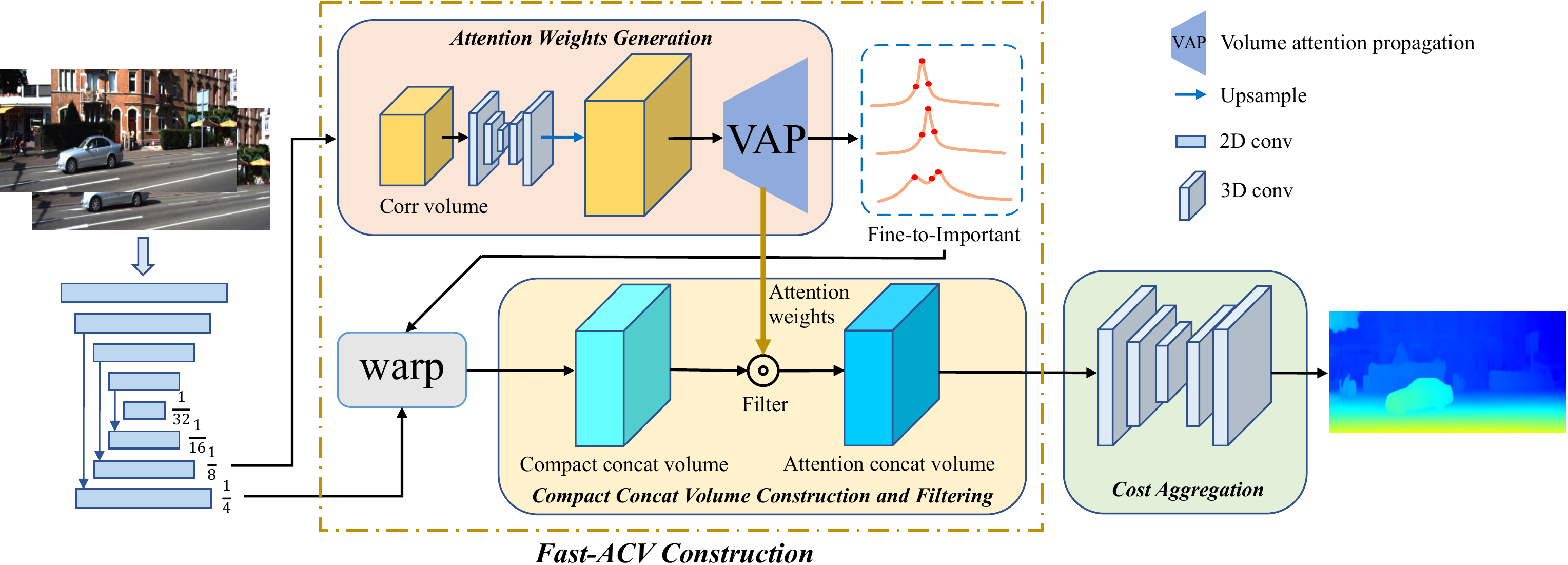}
    \caption{The structure of our proposed Fast-ACVNet. We first exploit a correlation volume to generate disparity hypotheses with high likelihood and the corresponding attention weights. Then we use the attention weights to filter the compact concatenation volume constructed based on disparity hypotheses, deriving our Fast-ACV.}
    \label{fig:fast_acvnet}
    \vspace{-10pt}
\end{figure*}

\subsection{Attention Concatenation Volume}\label{sec:acv}
The construction process of attention concatenation volume consists of three steps: attention weights generation, initial concatenation volume construction and attention filtering.

\vspace{2mm}
\noindent\textbf{1) Attention Weights Generation.}\label{sec:att_weighs} The attention weights aim to filter the initial concatenation volume so as to emphasize useful information and suppress irrelevant information. To this end, we generate attention weights by extracting geometric information from correlations between a pair of stereo images. Conventional correlation volume is obtained by computing pixel-to-pixel similarity which becomes unreliable for textureless regions due to lack of sufficient matching clues. To address this problem, we propose a more robust correlation volume construction method via multi-level adaptive patch matching (MAPM). We obtain feature maps at three different levels $l_1$, $l_2$ and $l_3$ from the feature extraction module, and the number of channels for $l_1$, $l_2$ and $l_3$ is 64, 128 and 128 respectively. For each pixel at a particular level, we utilize an atrous patch with a predefined size and adaptively learned weights to calculate the matching cost. By controlling the dilation rate, we ensure that the patch’s scope is related to the feature map level and meanwhile maintains the same number of pixels in similarity calculation for the center pixel. The similarity of two corresponding pixels is then a weighted sum of correlations between corresponding pixels within in the patch.

We split features into groups and compute correlation maps group by group~\cite{guo2019group}. Three levels of feature maps of $l_1$, $l_2$ and $l_3$ are concatenated to form $N_f$-channel unary feature maps ($N_f$=320). We equally divide $N_f$ channels into $N_g$ groups ($N_g$=40), and accordingly the first 8 groups are from $l_1$, the middle 16 groups are from $l_2$, and the last 16 groups are from $l_3$. Feature maps of different levels will not interfere with each other. We denote the $g^{th}$ feature group as $\mathbf{f}_l^g$, $\mathbf{f}_r^g$, and multi-level patch matching volume $\mathbf{C}_{patch}$ is computed as,
\begin{equation}
\begin{split}
\mathbf{C}_{patch}^{l_{k}}(g,d,x,y)=\frac{1}{N_f/N_g} \sum\limits_{(i,j)\in\Omega^k}\omega_{ij}^{k,g}\cdot C_{ij}^g(d,x,y) \\
C_{ij}^g(d,x,y)=\langle\mathbf{f}_l^g(x\!-\!i,y\!-\!j), \mathbf{f}_{r}^g(x\!-\!i-\!d,y\!-\!j)\rangle,
\end{split}
\end{equation}
where $\mathbf{C}_{patch}^{l_{k}}$ ($k$$\in$$(1,2,3)$) represents the matching cost of the feature level $k$, $\langle\cdot,\cdot\rangle$ is the inner product, $(x,y)$ represents the pixel's location, and $d$ denotes a disparity level. $\Omega^k$=$(i,j)$ ($i,j$$\in$$(-k,0,k)$) is a nine-point coordinate set, defining the scope of the patch on the $k$-level feature maps ($k$$\in$$(1,2,3)$). $\omega_{ij}^{k}$ represents the weight of a pixel $(i,j)$ in the patch on the $k$-level feature maps and is learned adaptively during the training process. The final multi-level patch matching volume is then obtained by concatenating matching costs $\mathbf{C}_{patch}^{l_k}$ ($k$$\in$$(1,2,3)$ of all levels,
\begin{equation}
\mathbf{C}_{patch}=\text{Concat}\left\{\mathbf{C}_{patch}^{l_1},\mathbf{C}_{patch}^{l_2},\mathbf{C}_{patch}^{l_3}\right\},
\end{equation}
we denote the derived multi-level patch matching volume as $\mathbf{C}_{patch}\in\mathbb{R}^{N_g\times{D}/4\times{H}/4\times{W}/4}$.
We then apply two 3D convolutions and a 3D hourglass network~\cite{guo2019group}
to regularize $\mathbf{C}_{patch}$, and then use another convolution layer to compress the channels to 1 and derive the attention weights, i.e. $\mathbf{A}\in\mathbb{R}^{1\times{D}/4\times{H}/4\times{W}/4}$.

To obtain accurate attention weights of different disparities to filter the initial concatenation volume, we use the ground truth disparity to supervise $\mathbf{A}$. Specifically, we use the $soft$ $argmin$ function~\cite{kendall2017end} to obtain the disparity estimation $\mathbf{d}_{att}$ from $\mathbf{A}$. We compute smooth L1 loss between $\mathbf{d}_{att}$ and the disparity ground truth to guide network learning.

\vspace{2mm}
\noindent\textbf{2) Initial Concatenation Volume Construction.} Given an input stereo image pair whose size is $H{\times}W{\times}3$, for each image, we obtain unary feature maps $\mathbf{f}_l$ and $\mathbf{f}_r$ for the left and right images respectively from CNN feature extraction. The size of feature maps of $\mathbf{f}_l$ ($\mathbf{f}_r$) is $N_c{\times}H/4{\times}W/4$. For initial concatenation volume construction, our implementation follows the design of PSMNet\cite{chang2018pyramid}, and thus we choose $N_c=32$ as PSMNet. The initial concatenation volume is formed by concatenating the $\mathbf{f}_l$ and $\mathbf{f}_r$ for each disparity level as,
\begin{equation}
\mathbf{C}_{concat}(\cdot,d,x,y)=\text{Concat}\left\{\mathbf{f}_{l}(x,y),\mathbf{f}_{r}(x-d,y)\right\},
\end{equation}
the accordingly size of $\mathbf{C}_{concat}$ is $2N_c\times{D}/4\times{H}/4\times{W}/4$, $D$ denotes the maximum of disparity.

\vspace{2mm}
\noindent\textbf{3) Attention Filtering.} After obtaining the attention weights $\mathbf{A}$, we use it to eliminate redundant information in the initial concatenation volume and in turn enhance its representation ability. The attention concatenation volume $\mathbf{C}_{ACV}$ at channel $i$ is computed as,
\begin{equation}
\mathbf{C}_{ACV}(i)=\mathbf{A}\odot\mathbf{C}_{concat}(i),
\label{equ:acv}
\end{equation}
where $\odot$ represents the element-wise product, and the attention weights $\mathbf{A}$ are applied to all channels of the initial concatenation volume.

\subsection{Fast Attention Concatenation Volume}\label{sec:fast_acv}

Compared with ACV, our Fast-ACV accelerates the process from two aspects. First, Fast-ACV employs a novel volume attention propagation (VAP) module which enables it to perform the majority of calculations for generating disparity hypotheses with high likelihood and their attention weights at a low resolution. Second, our Fast-ACV adopts a Fine-to-Important (F2I) sampling strategy to construct and filter a compact concatenation volume based on only the high likelihood disparity hypotheses to further reduce the time and memory cost. In the following, we present details of our VAP, compact concatenation volume construction via F2I and filtering.

\begin{figure}
\begin{center}
\includegraphics[width=1.0\linewidth]{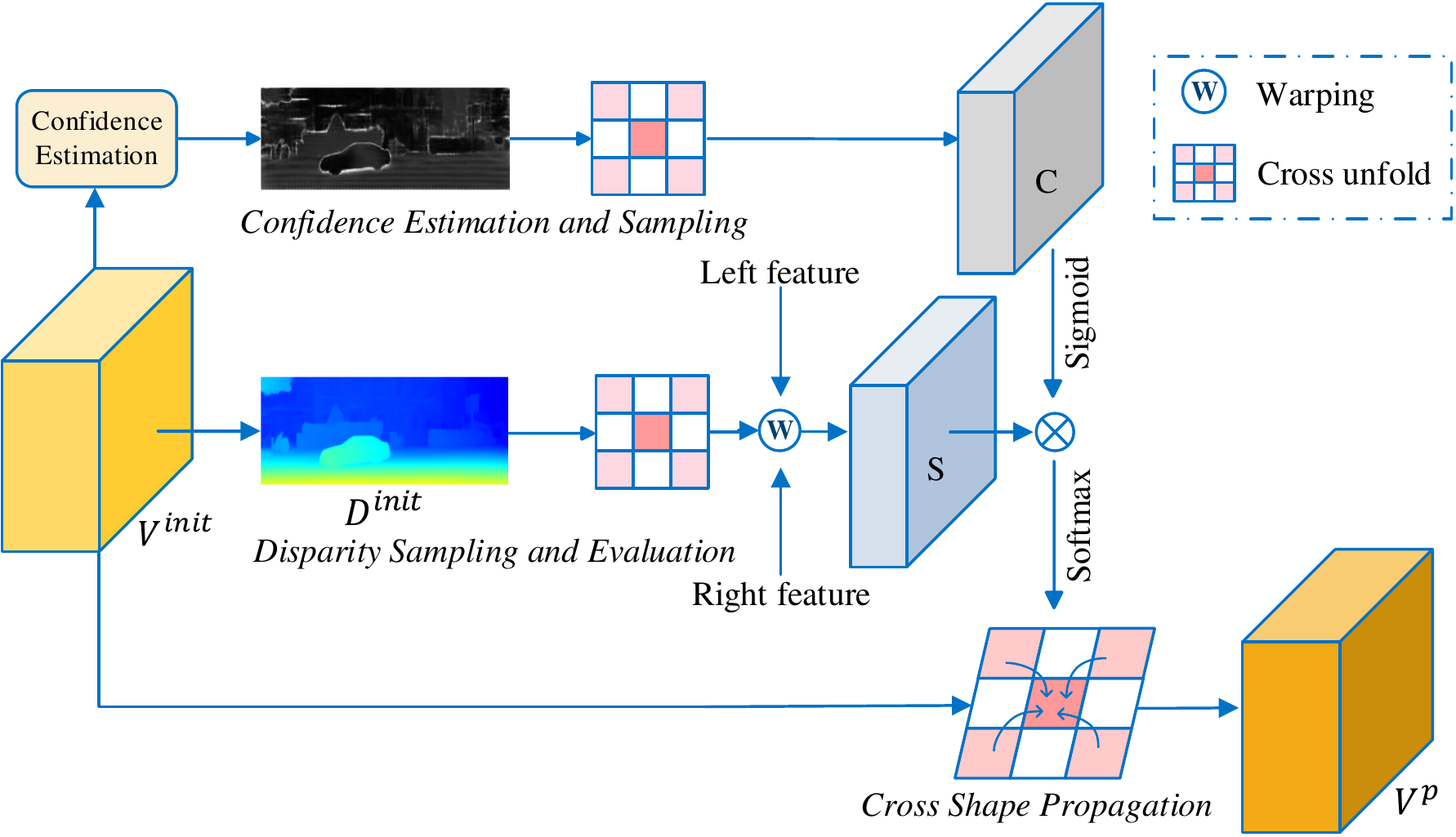}
\end{center}
\vspace{-10pt}
\caption{Volume Attention Propagation.}
\label{fig:vap}
\vspace{-15pt}
\end{figure}

\subsubsection{Volume Attention Propagation}
As shown in Fig. \ref{fig:fast_acvnet}, we construct and aggregate a 4D correlation volume at a low resolution and then form an initial high-resolution correlation volume via bilinear interpolation. In order to obtain a set of seed pixels that are likely to have accurate matches (i.e. a high accuracy for disparity prediction) and meanwhile have sharply distributed cost values along all the disparity levels (i.e. high confidence for disparity prediction), we conduct joint screening of neighboring disparity from two complementary perspectives: 1) \textbf{Disparity Sampling and Evaluation} to obtain matching scores, which represent the correlation degree of the surrounding disparities; 2) \textbf{Confidence Estimation and Sampling} to obtain confidence scores, which represent the degree of reliability of the surrounding disparities. Finally, the two evaluation scores are integrated through the \textbf{Cross Shape Propagation} which follows a principle that only the disparity with high correlation and high reliability simultaneously is the first choice for propagating to the current location. We detail each of the three components below.

\vspace{2mm}
\noindent\textbf{1) Disparity Sampling and Evaluation.} Given an initial high-resolution correlation volume $\mathbf{V}^{init}\in\mathbb{R}^{D/4 \times H/4 \times W/4}$ (as shown in Fig. \ref{fig:vap}) up-sampled from a low-resolution correlation volume via bilinear interpolation, we first regress an initial disparity map $\mathbf{D}^{init}\in\mathbb{R}^{H/4 \times W/4}$ from $\mathbf{V}^{init}$, and then sample disparities of adjacent pixels through convolution with a pre-defined one-hot filter pattern. To fully exploit neighboring information, the sampling scope is related to the up-sampling factor. For example, when the up-sampling factor is 2, we set the size of the sampling block to 3 $\times$ 3. In our experiments, we only perform a single round of iterative propagation which is sufficient to transfer reliable information from the seeds to the ambiguous boundary pixels. Meanwhile, for efficiency, we sample disparities around each pixel in a cross shape instead of a square shape. As shown in Fig. \ref{fig:vap}, the candidate disparities at pixel $i$ after sampling is $\mathbf{D}^{init}_{m}(i)$, $m=1,2,3,4,5$. Matching score at pixel $i$ is computed as,
\begin{equation}
\mathbf{S}_{m}(i)=\left<\mathbf{F}_{l}(i),\mathbf{F}_{r}\left(i-\mathbf{D}_{m}^{init}(i)\right)\right>, m=1,2,3,4,5,
\end{equation}
where $\left<\cdot,\cdot\right>$ is the inner product, $\mathbf{F}_l,\mathbf{F}_r \in\mathbb{R}^{C \times H/4 \times W/4}$ are left and right feature maps at 1/4 resolution from feature extraction and provide details for calculating matching scores for VAP to alleviate the blurred edges problem.

\begin{figure}
\begin{center}
\includegraphics[width=1.0\linewidth]{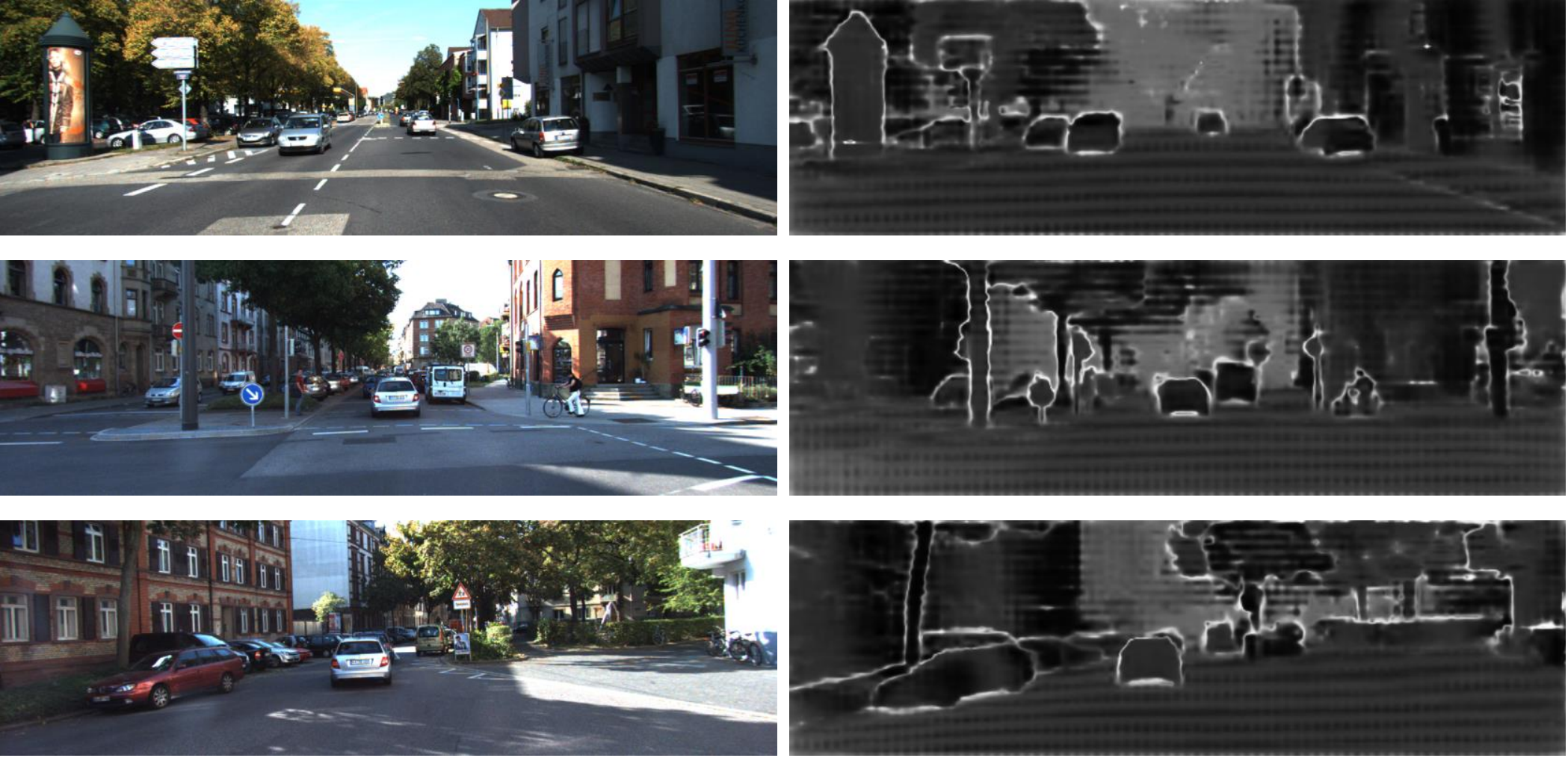}
\end{center}
\caption{Visualization of confidence maps. For edge regions, the disparity probability distribution is multi-peak due to interpolations from different objects, so the variance-based uncertainty is larger, which represents low confidence. The bright regions represent low confidence.}
\label{fig:confidence}
\vspace{-10pt}
\end{figure}

\vspace{2mm}
\noindent\textbf{2) Confidence Estimation and Sampling.} With the high-resolution correlation volume $\mathbf{V}^{init}$, we obtain a probability volume $\mathbf{P}^{init}\in\mathbb{R}^{D/4 \times H/4 \times W/4}$ by $softmax$. Due to the disparity discontinuity in the edge regions, the problem of multi-peak distribution is easily caused by linear interpolation in the edge regions, as shown in Fig. \ref{fig:confidence}, which causes low confidence and high prediction error. Thus, pixels with multi-peak distribution are unreliable and should not be used for propagation. To this end, we propose to employ uncertainty estimation to evaluate the pixel-level confidence of the current estimation. The uncertainty is estimated by the variance of the distribution. The uncertainty $\mathbf{U}(i)$ at pixel $i$ is calculated as:
\begin{equation}
\mathbf{U}(i)=\sum\limits_{d=0}^{D/4-1} \mathbf{P}^{init}_{d}(i) \times \left(d-\mathbf{D}^{init}(i)\right)^{2},
\end{equation}
where $\mathbf{P}^{init}_{d}(i)$ means the $d^{th}$ probability value at pixel $i$. With uncertainty $\mathbf{U}(i)$, the confidence at pixel $i$ is defined as,
\begin{equation}
\mathbf{C}(i)=\alpha + \beta \times \mathbf{U}(i),
\end{equation}
where $\alpha, \beta$ are learned parameters. The greater the uncertainty is, the lower the confidence will be. For the pixel $i$, we sample confidence around it in a cross shape to obtain the candidate confidence scores which are $\mathbf{C}_{m}(i)$, $m=1,2,3,4,5$, as shown in Fig. \ref{fig:vap}.

\vspace{2mm}
\noindent\textbf{3) Cross Shape Propagation.} Combining matching score and confidence score of each neighboring pixel, we can obtain an overall propagation weight as,
\begin{equation}
\begin{aligned}
\mathbf{W}_{m}(i) = \mathbf{S}_{m}(i) \times sigmoid\left(\mathbf{C}_{m}(i) \right).
\end{aligned}
\end{equation}
Along the spatial dimension, we unfold the initial correlation volume $\mathbf{V}^{init}$ in the cross shape, and then obtain the unfolded correlation volume $\mathbf{V}^{u} \in \mathbb{R}^{M \times D/4 \times H/4 \times W/4}$. At spatial position $i$ and disparity $d$ of $\mathbf{V}^{u}$, we can obtain the vector $\mathbf{V}^{u}(i,d) \in \mathbb{R}^{M}$. The cross shape propagation operation is defined as,
\begin{equation}
\begin{aligned}
\mathbf{V}^{p}(i,d) = \sum\limits_{m=1}^{M} \mathbf{V}_{m}^{u}(i,d) \times softmax(\mathbf{W}_{m}(i)),
\end{aligned}
\end{equation}
$\mathbf{V}^{p} \in \mathbb{R}^{D/4 \times H/4 \times W/4}$ is the correlation volume after propagation (see Fig. \ref{fig:vap}).

\begin{figure*}[t]
    \centering
    \includegraphics[width=1.0\linewidth]{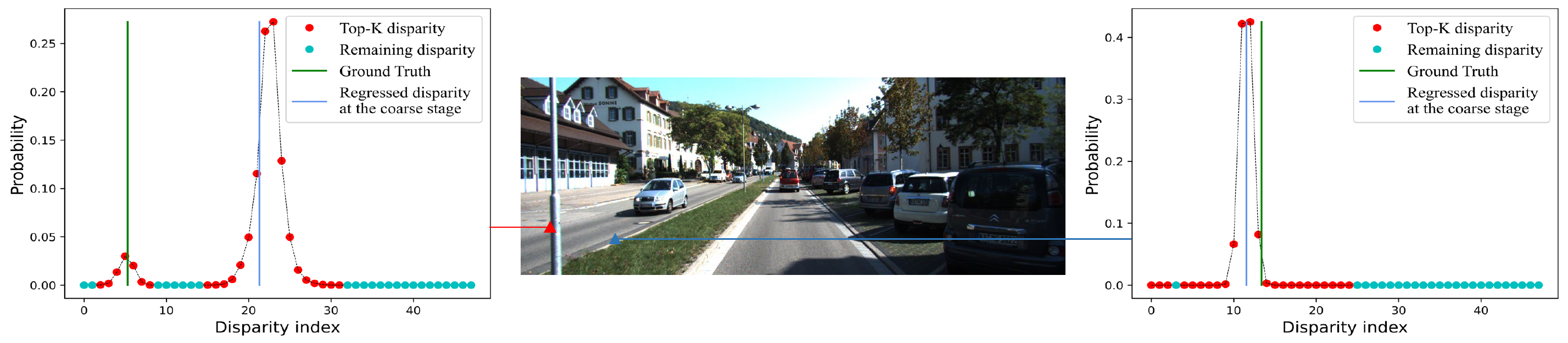}
    \caption{Visualization of the Fine-to-Important sampling strategy (Top-K). Ideally, the disparity probability distribution should be unimodal peaked at true disparities. However, the actual probability distribution could be either unimodal or multimodal at some pixels. The red and blue triangles in the middle figure show two points with multimodal (left figure) and unimodal distributions (right figure) at the coarse stage respectively. The cascaded methods only sample the disparities near the regressed disparity of the coarse stage (denoted by blue lines), however, which could miss the true disparity for points with multimodal distributions. In contrast, Our sampling strategy (red dots) can preserve true disparities for points with both multimodal and unimodal distributions.}
    \label{fig:topk_sample}
    \vspace{-10pt}
\end{figure*}

\subsubsection{Compact Concatenation Volume Construction via F2I and Filtering.}
For Fast-ACV, we propose a Fine-to-Important (F2I) sampling strategy to preserve high likelihood hypotheses based on which we construct a compact concatenation volume and their corresponding attention weights for filtering the compact concatenation volume. Such strategy can greatly reduce computational cost with little accuracy degradation. With the correlation volume $\mathbf{V}^{p}$, we obtain the disparity probability distribution volume $\mathbf{P}\in\mathbb{R}^{D/4 \times H/4 \times W/4}$ by applying the $softmax$ function to it. Ideally, the disparity probability distribution $\mathbf{P}$ should be unimodally peaked at true disparities. However, the actual probability distribution could be multimodal at some pixels, such as ill-posed areas, texture-less regions, and occlusions (see Fig. \ref{fig:topk_sample}). To concurrently preserve high likelihood hypotheses and reduce the computational cost, we utilize the F2I sampling strategy which selects the top K probability values of $\mathbf{P}$ at every pixel as attention weights $\mathbf{A}^{F}\in\mathbb{R}^{K \times H/4 \times W/4}$, and use their corresponding disparities as high likelihood hypotheses $\mathbf{D}^{hyp}\in \mathbb{R}^{K \times H/4 \times W/4}$,
\begin{equation}
\begin{aligned}
\mathbf{A}^{F} = max^{K}_{i=1}\{\mathbf{P}\},
\end{aligned}
\end{equation}
\begin{equation}
\begin{aligned}
\mathbf{D}^{hyp} = arg\; max^{K}_{i=1}\{\mathbf{P}\},
\end{aligned}
\end{equation}

The compact concatenation volume is then constructed
by concatenating the $\mathbf{f}_l^{F}$ and $\mathbf{f}_r^{F}$ based on only the high likelihood disparity hypotheses as,
\begin{equation}
\mathbf{C}_{concat}^{compact}(\cdot,\mathbf{D}^{hyp}_{xy},x,y)\!=\!\text{Concat}\left\{\mathbf{f}_{l}^{F}(x,y),\mathbf{f}_{r}^{F}(x\!-\!\mathbf{D}^{hyp}_{xy},y)\right\}
\end{equation}
The unary feature maps $\mathbf{f}_l^{F}$ and $\mathbf{f}_r^{F}$ for the left and right images respectively is from CNN feature extraction. Finally, we use attention weights $\mathbf{A}^{F}$ to filter the compact concatenation volume to enhance its representation ability. The fast attention concatenation volume $\mathbf{C}_{F\!-\!ACV}$ at channel $i$ is computed as,
\begin{equation}
\mathbf{C}_{F\!-\!ACV}(i)=\mathbf{A}^{F}\odot\mathbf{C}_{concat}^{compact}(i).
\label{equ:fast_acv}
\end{equation}

\subsection{ACVNet Architecture} \label{sec:acvnet_architecture}
Based on the ACV, we design an accurate and efficient end-to-end stereo matching network, named ACVNet. Fig. \ref{fig:acvnet} shows the architecture of our ACVNet which consists of four steps of Feature Extraction, Attention Concatenation Volume Construction, Cost Aggregation and Disparity Prediction. In the following, we introduce each step in detail.

\vspace{2mm}
\noindent\textbf{Feature Extraction.} We adopt the three-level ResNet-like architecture in~\cite{guo2019group}. We first use a set of convolutions of $3{\times}3$ kernel and 32 channels to downsample the input images. Then, 16 residual layers~\cite{he2016deep}  with 64 channels are followed to produce unary features at 1/4 resolution, i.e., $l_1$. After that, we apply 6 residual layers with 128 channels to obtain large receptive fields and semantic information, i.e., $l_2$ and $l_3$. Finally, all feature maps ($l_1$, $l_2$, $l_3$) at 1/4 resolution are concatenated to form 320-channel feature maps for the generation of attention weights. Then two convolutions are applied to compress the 320-channel feature maps to 32-channel feature maps for construction of the initial concatenation volume, which are denoted as $\mathbf{f}_l$ and $\mathbf{f}_r$. We utilize GwcNet\cite{guo2019group} as the baseline model to implement ACVNet.  Thus, we follow GwcNet to use 64, 128, and 128 channels for $l_1$, $l_2$, and $l_3$ respectively.

\vspace{2mm}
\noindent\textbf{Attention Concatenation Volume Construction.} We take the 320-channels feature maps for attention weights generation, and $\mathbf{f}_l$ and $\mathbf{f}_r$ for initial concatenation volume construction. Then attention weights are used to filter the initial concatenation volume to produce a 4D cost volume for all disparities, as described in Sec. \ref{sec:acv}.

\vspace{2mm}
\noindent\textbf{Cost Aggregation.} We process the ACV using a pre-hourglass module which consists of four 3D convolutions with batch normalization and ReLU, and two stacked 3D hourglass networks~\cite{guo2019group}. Each hourglass network consists of four 3D convolutions and two 3D deconvolutions stacked in an encoder-decoder architecture. 

\vspace{2mm}
\noindent\textbf{Disparity Prediction.} We obtain three outputs from cost aggregation. For each output, following GwcNet~\cite{guo2019group}, we convolve it using two 3D convolutions to output a 1-channel 4D volume. Then we up-sample and convert it into a probability volume by the $softmax$ function along the disparity dimension. Finally, the predicted value is computed by the $soft$ $argmin$ function~\cite{kendall2017end}.
The three predicted disparity maps are denoted as $\mathbf{d}_0$, $\mathbf{d}_1$, $\mathbf{d}_2$.

\subsection{Fast-ACVNet Architecture} \label{sec:fast_acvnet_architecture}
We construct a real-time stereo model, named Fast-ACVNet, based on our Fast-ACV. It is worth explaining that Fast-ACVNet is different from ACVNet-Fast in our CVPR 2022 version\cite{acvnet}. Compared with ACVNet-Fast, Fast-ACVNet utilizes the newly proposed VAP and Fine-to-Important sampling strategy to achieve a real-time speed and state-of-the-art accuracy among the efficiency-oriented methods. Fig. \ref{fig:fast_acvnet} shows the architecture of our Fast-ACVNet which consists of four steps of Multi-scale Feature Extraction, Fast Attention Concatenation Volume Construction, Cost Aggregation and Disparity Prediction. 

\vspace{2mm}
\noindent\textbf{Multi-scale Feature Extraction.} Given an input stereo image pair whose size is $H{\times}W{\times}3$, we use the MobileNetV2 pre-trained on ImageNet~\cite{deng2009imagenet} to obtain four scales of feature maps whose resolutions are 1/4, 1/8, 1/16, and 1/32 of the original resolution respectively. Then we use three up-sampling blocks with skip-connections to increase the size of low-resolution feature maps of $H/32{\times}W/32$, $H/16{\times}W/16$ and $H/8{\times}W/8$ resolution (see Fig.\ref{fig:fast_acvnet}). 
Finally, we obtain $H/8{\times}W/8$ resolution feature maps for generation of attention weights, and $H/4{\times}W/4$ resolution feature maps for construction of the compact concatenation volume.

\vspace{2mm}
\noindent\textbf{Fast Attention Concatenation Volume Construction.} We take 96-channels feature maps at 1/8 resolution for attention weights generation, and $\mathbf{f}_l^{F}$ and $\mathbf{f}_r^{F}$ at 1/4 resolution for compact concatenation volume construction. We divide the 96 channels of feature maps at 1/8 resolution into 12 groups, each group contains 8 channels of feature maps. Then we construct a group-wise correlation volume as~\cite{guo2019group}. We utilize a lightweight guided hourglass network same as CoEx~\cite{bangunharcana2021correlate} to regularize the low-resolution group-wise correlation volume. Then we apply VAP to the regularized correlation volume to obtain the volume $\mathbf{V}^{p}$ at 1/4 resolution. We further convert $\mathbf{V}^{p}$ into a probability volume to get the attention weights $\mathbf{A}^{F}\in\mathbb{R}^{K \times H/4 \times W/4}$ by using the Top-K probability values of $\mathbf{V}^{p}$ at every pixel. The attention weights are used to filter the compact concatenation volume to produce Fast-ACV, as described in Sec. \ref{sec:fast_acv}.

\vspace{2mm}
\noindent\textbf{Cost Aggregation.} We aggregate the Fast-ACV using a guided hourglass network, which consists of six 3D convolutions and two 3D deconvolutions stacked in an encoder-decoder architecture. Following CoEx~\cite{bangunharcana2021correlate}, we utilize extracted feature maps from the left image as guidance for cost aggregation.

\vspace{2mm}
\noindent\textbf{Disparity Prediction.} We take out only the top 2 values at every pixel of the aggregated cost volume and perform $softmax$ on these values to compute the expected disparity values. Finally, we up-sample the output disparity prediction to the original input image resolution by "superpixel" weights surrounding each pixel as~\cite{yang2020superpixel}.

\subsection{Loss Function} \label{sec:loss}
For ACVNet, the final loss is given by,
\begin{equation}
\begin{split}
	L=\lambda_{att}\cdot\text{Smooth}_{L_1}(\mathbf{{d}}_{att}-\mathbf{d}^{gt})+\\ \sum_{i=0}^{i=2}\lambda_i\cdot\text{Smooth}_{L_1}(\mathbf{d}_i-\mathbf{d}^{gt}),
\end{split}
\label{equ:loss}
\end{equation}
where $\mathbf{d}_{att}$ is obtained by attention weights $\mathbf{A}$ in Sec. \ref{sec:acv}. $\lambda_{att}$ represents the coefficient for the predicted $\mathbf{{d}}_{att}$,  $\lambda_i$ represents the coefficient for the $i^{th}$ predicted disparity and $\mathbf{d}^{gt}$ denotes the ground-truth disparity map. The $\text{Smooth}_{L_1}$ is the smooth L1 loss.

For Fast-ACVNet, the final loss is given by,
\begin{equation}
	L^{F}=\lambda_{att}^F\cdot\text{Smooth}_{L_1}(\mathbf{{d}}_{att}^F-\mathbf{d}^{gt})+\lambda^F\cdot\text{Smooth}_{L_1}(\mathbf{{d}}^F-\mathbf{d}^{gt}),
\label{equ:loss}
\end{equation}
where $\mathbf{d}_{att}^F$ is obtained by compact attention weights for Fast-ACV, $\mathbf{d}^F$ is final output of Fast-ACVNet.

\begin{table*}
\centering
\caption{Ablation study of the ACV on Scene Flow~\cite{dispNetC2016large}.} \label{tab:acv}
\begin{tabular}{lccccccccccc}
\hline
\multirow{2}{*}{Model} & Attention   & Hourglass & Supervise & \textgreater1px & \textgreater2px &\textgreater3px & D1 & EPE & FLOPs & Params.\\ 
& Weights & for Attention & for Attention & (\%) & (\%)& (\%) & (\%) & (px) & (G) &(M) \\
\hline
GwcNet~\cite{guo2019group} & & & &8.03 &4.47 &3.30 & 2.71 & 0.76 &976.4 &6.91 \\
Gwc-att  & \checkmark & & &6.14 &3.39 &2.49& 2.03 & 0.57 &982.1 &7.02\\
Gwc-att-hg & \checkmark & \checkmark & &5.67 &3.09 &2.23& 1.87 & 0.52 & 1071.9 & 7.40 \\
Gwc-att-hg-s & \checkmark & \checkmark & \checkmark & 4.89 &2.69 &1.98 & 1.55 & 0.46 & 1071.9 & 7.40 \\
\hline
\end{tabular}
\end{table*}

\section{Experiment} \label{sec:experiment}
\subsection{Datasets and Evaluation Metrics} \label{sec:Datasets}
\noindent\textbf{Scene Flow} is a collection of synthetic stereo datasets which provides 35,454 training image pairs and 4,370 testing image pairs with a resolution of 960$\times$540. This dataset provides dense disparity maps as ground truth. For Scene Flow~\cite{dispNetC2016large} dataset, we utilized the widely-used evaluation metrics end-point error (EPE) and percentage of disparity outliers D1 as the evaluation metrics. The outliers are defined as the pixels whose disparity errors are greater than $\text{max}(3\text{px}, 0.05d^{*})$, where $d^{*}$ denotes the ground-truth disparity. 

\noindent\textbf{KITTI} includes KITTI 2012~\cite{kitti2012} and KITTI 2015~\cite{kitti2015}, which are datasets for real-world driving scenes. KITTI 2012 contains 194 training pairs and 195 testing pairs, and KITTI 2015 contains 200 training pairs and 200 testing pairs. Both datasets provide sparse ground-truth disparities obtained with LIDAR. 
For KITTI 2012, we report the percentage of pixels with errors larger than x disparities in both non-occluded (x-noc) and all regions (x-all), as well as the overall EPE in both non-occluded (EPE-noc) and all the pixels (EPE-all). For KITTI 2015, we report the percentage of pixels with EPE larger than 3 pixels in background regions (D1-bg), foreground regions (D1-fg), and all (D1-all). 

\noindent\textbf{ETH3D}~\cite{schops2017multi} is a collection of grayscale stereo pairs from indoor and outdoor scenes. It contains 27 training and 20 testing image pairs with sparse labeled ground-truth. Its disparity range is between 0 and 64. The percentage of pixels with errors larger than 2 pixels (bad 2.0) and 1 pixel (bad 1.0) are reported.

\noindent\textbf{Middlebury}~\cite{scharstein2014high} is an indoor dataset with 15 training image pairs and 15 testing image pairs with full, half, and quarter resolutions. Bad 2.0 (percentage of the points with absolute error larger than 2 pixels) are reported.

\subsection{Implementation Details} \label{sec:imple_detail}
We implement our methods with PyTorch and perform our experiments using NVIDIA RTX 3090 GPUs. For all the experiments, we use the Adam~\cite{kingma2014adam} optimizer, with $\beta_1=0.9$, $\beta_2=0.999$. For ACVNet, the coefficients of four outputs are set as $\lambda_{att}$=0.5, $\lambda_{0}$=0.5, $\lambda_{1}$=0.7, $\lambda_{2}$=1.0. For Fast-ACVNet, the coefficients of two outputs are set as $\lambda_{att}^F$=0.5, $\lambda^{F}$=1.0. For ACVNet on Scene Flow, we first train attention weights generation network for 64 epochs and then train the remaining network for another 64 epochs. Finally, we train complete network for 64 epochs. The initial learning rate is set to 0.001 decayed by a factor of 2 after epochs 20, 32, 40, 48, and 56. For Fast-ACVNet on Scene Flow, we first train attention weights generation network for 24 epochs, and then train complete network for another 24 epochs. The initial learning rate is set to 0.001 decayed by a factor of 2 after epochs 10, 15, 18, and 21. For KITTI, we finetune the pre-trained Scene Flow model on the mixed KITTI 2012 and KITTI 2015 training sets for 500 epochs. The initial learning rate is 0.001 and decreases to 0.0001 half at the 300th epoch.

\begin{figure*}
\centering
\includegraphics[width=1.0\textwidth]{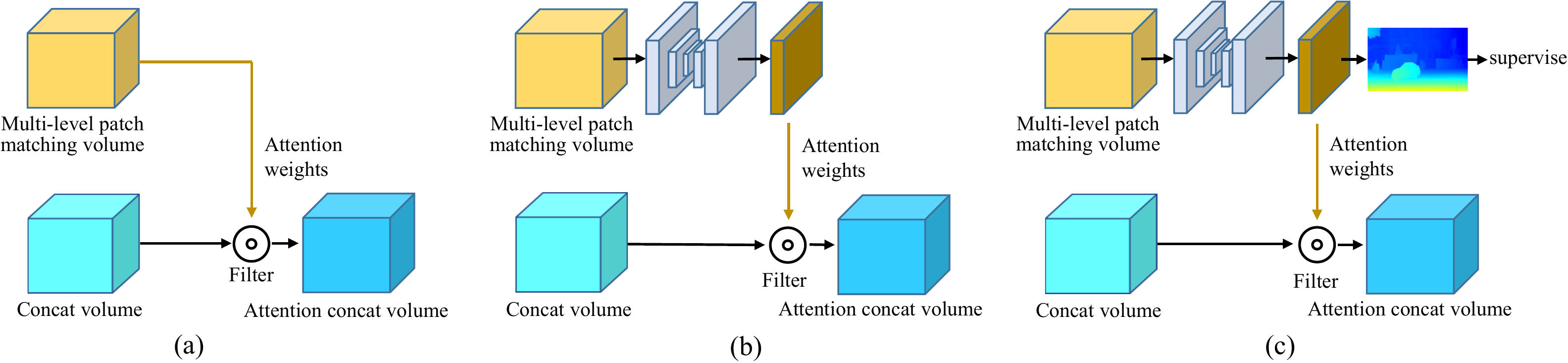} 
\caption{Illustration of different ways for constructing ACV.}
\label{fig:three_acv}
\vspace{-5pt}
\end{figure*} 

\begin{table} 
\begin{center}
\caption{Ablation study of VAP and F2I sampling strategy of Fast-ACV on Scene Flow.}\label{tab:vap_topk}
\begin{tabular}{lccccc}
\hline
\multirow{2}{*}{Base Model} & \multirow{2}{*}{VAP} & \multirow{2}{*}{F2I} & D1 & EPE & Runtime \\ 
 & &  & (\%) & (px) & (ms)\\ 
\hline
\makecell{Fast-ACVNet \\ (baseline)} &  &48  & 2.51 & 0.66 & 49\\
\hline
& \checkmark & 48 & 2.32 & 0.62 & 50\\
& \checkmark & 32 & 2.40 & 0.63 & 43\\
& \checkmark & 24 & 2.49 & 0.64  & 39\\
& \checkmark & 16 & 2.60 & 0.69 & 35\\
\hline
\end{tabular}
\end{center}
\vspace{-10pt}
\end{table}

\begin{table} 
\begin{center}
\caption{Ablation study of Fast-ACV on Scene Flow. Lightweight denotes using the lightweight backbone and aggregation network.}\label{tab:acv_to_fastacv}

\begin{tabular}{lccccc}
\hline
\multirow{2}{*}{Model} &\multirow{2}{*}{Lightweight} &\multirow{2}{*}{VAP} & \multirow{2}{*}{F2I (24)} & EPE & Runtime \\ 
 & &  & & (px) & (ms)\\ 
\hline
ACVNet & & & & 0.48 & 200 \\
ACVNet &\checkmark & & & 0.59 & 62 \\ 
ACVNet &\checkmark & \checkmark & & 0.62 & 50 \\ 
Fast-ACVNet &\checkmark &\checkmark &\checkmark & 0.64 & 39 \\
\hline
\end{tabular}
\end{center}
\vspace{-5pt}
\end{table}

\begin{table} 
\begin{center}
\caption{Performance of ACVNet when using different number of parameters in aggregation on Scene Flow~\cite{dispNetC2016large}} \label{tab:acv_performance}
\begin{tabular}{cccccc}
\hline
Model & ACV & \makecell{Hourglass \\ Number} & \makecell{D1\\(\%)} & \makecell{EPE\\(px)} & \makecell{Params.\\(M)} \\ 
\hline
GwcNet~\cite{guo2019group} &  & 3 & 2.71 & 0.76 & 6.91 \\
Gwc-acv-3 & \checkmark & 3 & 1.55 & 0.46 & 7.40 \\
Gwc-acv-1 & \checkmark & 1 & 1.79 & 0.53 & 5.04 \\
Gwc-acv-0 & \checkmark & 0 & 2.08 & 0.59 & 3.86 \\
\hline
\makecell{ACVNet\\(Gwc-acv-2)} & \checkmark & 2 & 1.59 & 0.48 & 6.22 \\
\hline
\end{tabular}
\end{center}
\vspace{-10pt}
\end{table}

\begin{table} 
\begin{center}
\caption{Performance of Fast-ACVNet when using different number of parameters in aggregation on Scene Flow. It is noteworthy that Fast-ACV without using the attention filter is the cascaded cost volume. Our Fast-ACV outperforms cascaded cost volume by a large margin.}\label{tab:fast_acv_performance}
\begin{tabular}{c|ccccc}
\hline
Model & \makecell{Attention \\ Filter} & \makecell{Hourglass \\ Number} & \makecell{D1\\(\%)} & \makecell{EPE\\(px)} & \makecell{Params.\\(M)} \\ 
\hline
\multirow{6}{*}{Fast-ACV} &  & 1 & 0.83 & 3.56 & 39 \\
&  & 2 & 0.78 & 3.23 & 46\\
&  & 3 & 0.74 & 2.92 & 52\\
\cline{2-6}
& \checkmark & 1 & 0.64 & 2.49 & 39\\
& \checkmark & 2 & 0.62 & 2.33 & 46\\
& \checkmark & 3 & 0.60 & 2.20 & 52\\
\hline
\end{tabular}
\end{center}
\end{table}

\begin{table*}
\centering
\caption{Comparisons with the original cost volume and cascade cost volume. \textbf{Bold}: Best.} \label{tab:vs_cascade}
\begin{tabular}{lccccccc}
\hline
{Method} & \textgreater1px (\%) & \textgreater2px (\%) &\textgreater3px (\%)& EPE (px) & FLOPs (G) &Params. (M) &Runtime (ms)\\
\hline
PSMNet~\cite{chang2018pyramid} &9.46 &5.19 & 3.80 & 0.88 & 961.9 & 5.22 & 310\\
PSM+CAS~\cite{gu2020cascade} &7.44 &4.61 & 3.50 & 0.72 & 1134.9 & 10.3 & 200\\
PSM+ACV &\textbf{7.35} &\textbf{4.12} & \textbf{3.01} & \textbf{0.63} & {1052.4} & {5.53} & {355}\\
PSM+Fast-ACV &{7.41} &{4.22} & {3.10} & {0.65} & \textbf{436.8} & \textbf{4.52} & \textbf{85}\\
\hline
GwcNet~\cite{guo2019group} &8.03 &4.47 & 3.30 & 0.76 & 976.4 & 6.91 & 180\\
Gwc+CAS~\cite{gu2020cascade} &7.46 &4.16 & 3.04 & 0.64 & 1248.8 & 10.7 & 220\\
Gwc+ACV &\textbf{4.89} &\textbf{2.69} & \textbf{1.98} & \textbf{0.46} & {1071.9} & {7.40} & {200}\\
Gwc+Fast-ACV &{7.09} &{3.96} & {2.87} & {0.61} & \textbf{456.5} & \textbf{5.35} & \textbf{81}\\
\hline
\end{tabular}
\end{table*}

\begin{table} 
\begin{center}
\caption{Evaluating the effectiveness of VAP by integrating it into other two stereo models on Scene Flow. The image size is 960$\times$512.}\label{tab:vap}
\begin{tabular}{lccc}
\hline
Model &\textgreater3px (\%) & EPE (px)  &FLOPs (G)\\ 
\hline
PSMNet~\cite{chang2018pyramid}  & 3.80 & 0.88 & 961.9\\
PSMNet-VAP  & \textbf{3.48} & \textbf{0.71} & 964.2\\
\hline
GwcNet~\cite{guo2019group}  & 3.30 & 0.76 & 976.4\\
GwcNet-VAP  & \textbf{2.52} & \textbf{0.59} & 979.2\\
\hline
\end{tabular}
\end{center}
\vspace{-10pt}
\end{table}

\begin{figure}
\begin{center}
\includegraphics[width=1.0\linewidth]{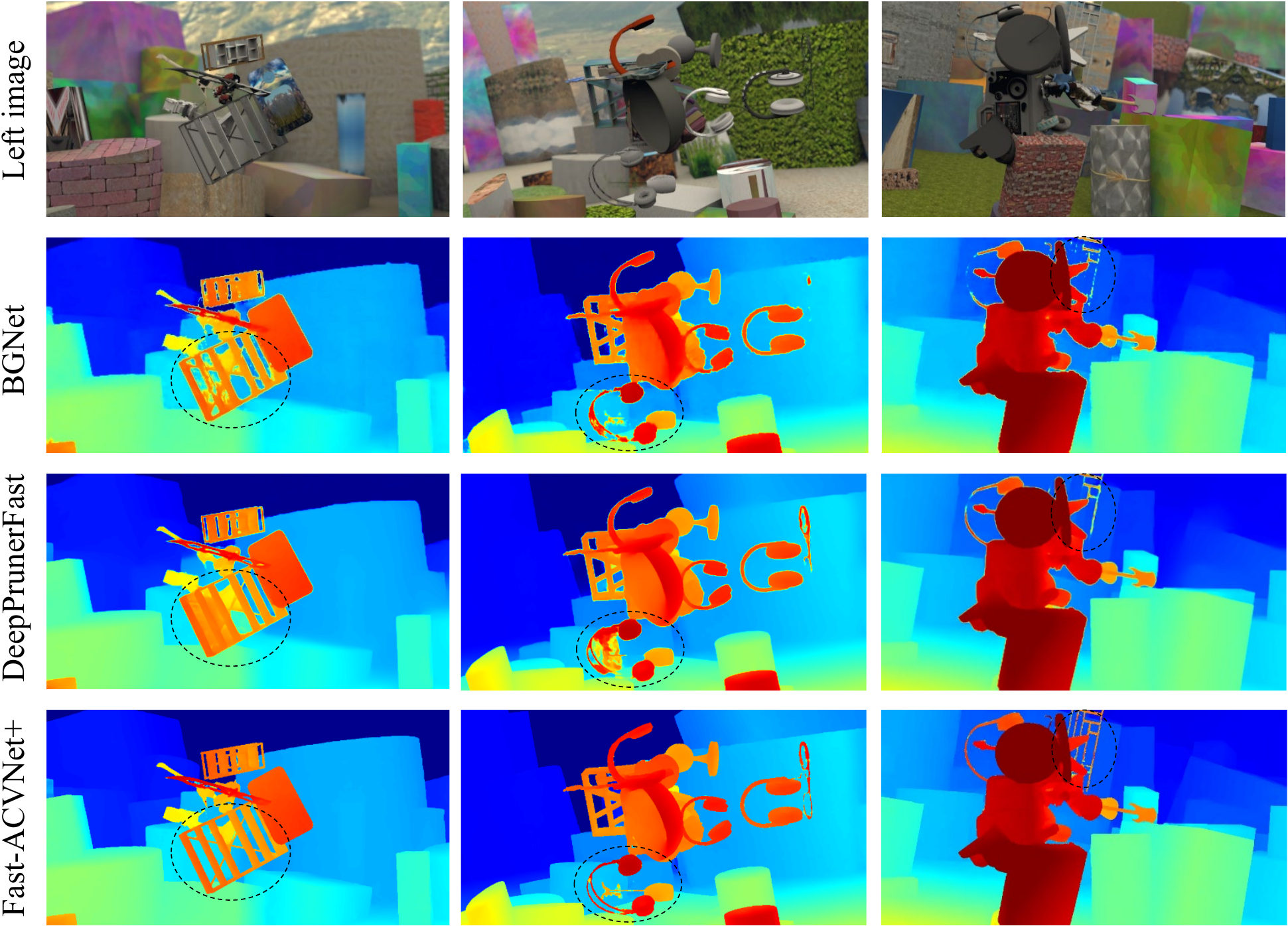} 
\end{center}
\vspace{-5pt}
\caption{Qualitative results on Scene Flow.}
\label{fig:sceneflow}
\vspace{-10pt}
\end{figure}

\begin{figure}
\centering
\includegraphics[width=1.0\linewidth]{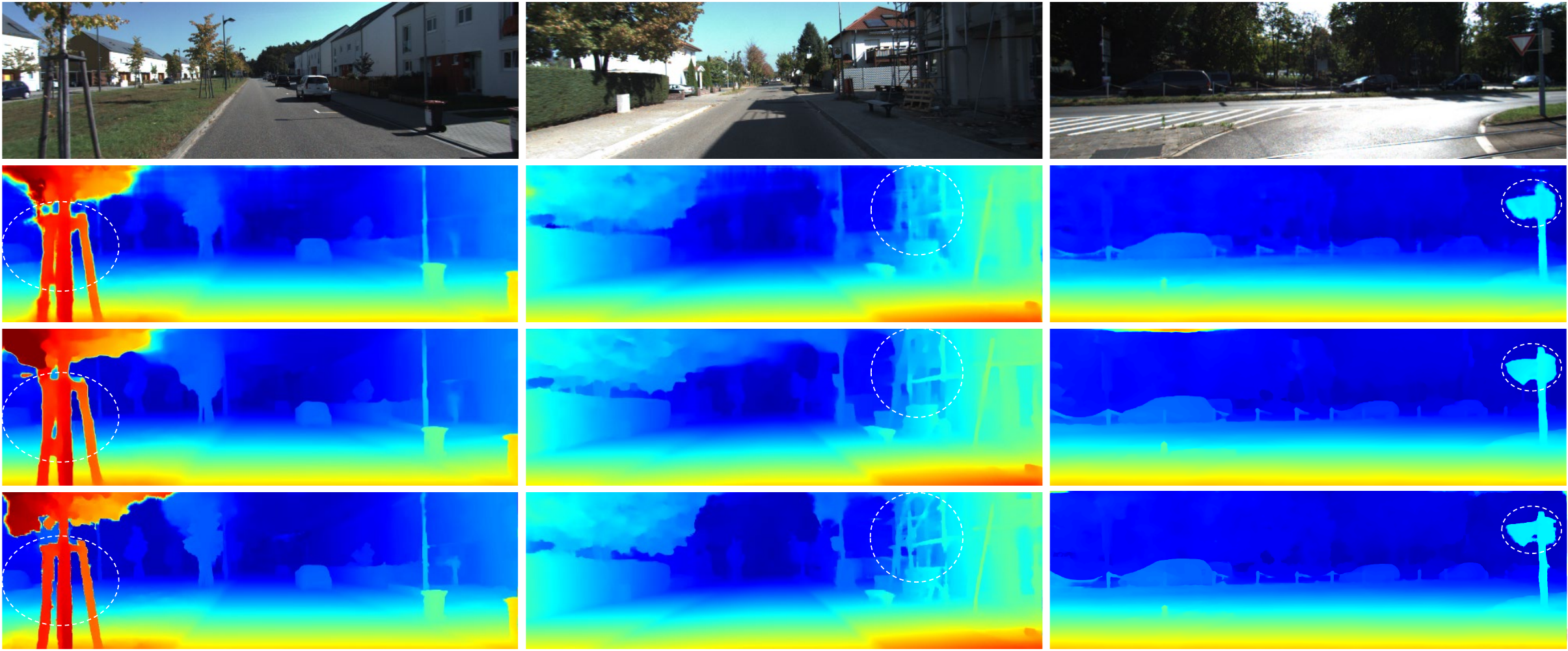} 
\caption{Qualitative results on KITTI 2012 and KITTI 2015. The second, third, and fourth row shows the results of BGNet, DeepPrunerFast and Fast-ACVNet+ respectively.}
\label{fig:kitti}
\end{figure}

\subsection{Ablation Study} \label{sec:ablation}
In this section, we investigate different designs and configuration settings for ACV and Fast-ACV.
\subsubsection{Attention Concatenation Volume.} \label{sec:acv_exp}
We evaluate different strategies for constructing the ACV on Scene Flow~\cite{dispNetC2016large}. We take GwcNet~\cite{guo2019group} for example and for other existing stereo matching methods based on cost volumes, the trend is similar and thus is omitted in this experiment. We replace the combined volume of GwcNet with our ACV and keep the subsequent aggregation and disparity prediction modules the same. Fig. \ref{fig:three_acv} shows three different ways of constructing an ACV. Fig. \ref{fig:three_acv} (a) directly averages the multi-level patch matching volume along the channel dimension and multiplies it with the concatenation volume, denoted as Gwc-att. As shown in Tab. \ref{tab:acv}, just this simple approach can dramatically improve the accuracy of GwcNet. Apparently, when using the correlation volume to filter the concatenation volume, the accuracy of correlation volume is crucial and largely affects the final performance of the network, so we use an hourglass architecture of 3D convolutions to aggregate it, which is denoted as Gwc-att-hg shown in Fig. \ref{fig:three_acv} (b). The results in Tab. \ref{tab:acv} show that Gwc-att-hg improves D1 and EPE by 9.8\% and 8.7\% respectively compared with Gwc-att. To further explicitly constrain the correlation volume during training, we use the $softmax$ and $soft$ $argmin$ functions for regression to obtain the predicted disparity and use the ground truth to supervise the disparity, denoted as Gwc-att-hg-s shown in Fig. \ref{fig:three_acv} (c). Compared with the Gwc-att-hg, Gwc-att-hg-s improves D1 and EPE by 17.1\% and 11.5\% respectively with no computational cost increase in the inference stage. Overall, by replacing the combined volume in GwcNet with our ACV, our Gwc-att-hg-s model achieves 42.8\% and 39.5\% improvement for D1 and EPE compared with GwcNet, demonstrating the effectiveness of ACV. As shown in Tab. \ref{tab:acv_performance}, we choose Gwc-acv-2 which utilizes two hourglass aggregation networks as our final model, and we denote it as ACVNet.

\subsubsection{Fast Attention Concatenation Volume.} \label{sec:fast_acv_exp}
 We first evaluate the effectiveness of VAP and different number of seed candidates in Fast-ACVNet, as shown in Tab. \ref{tab:vap_topk}. In this ablation study, we use Fast-ACVNet without using VAP but using only bilinear upsample to increase the resolution of a correlation volume as our baseline. In the baseline model, we set the number of values sampled in the correlation volume $\textbf{V}^{p}$ at each pixel as 48 (i.e. all disparity candidates at 1/4 resolution).  Results in Tab. \ref{tab:vap_topk} show that our VAP can improve EPE from 0.66 to 0.62. We further explore the influence of the sampling number of F2I for the final accuracy and runtime. As shown in Tab. \ref{tab:vap_topk}, we find that F2I, i.e., sample 24 candidate seeds, can achieve the best results in terms of accuracy and runtime. When the sampling value in F2I increases from 24 to 32 or 48, there is little increase in accuracy, but a large increase in runtime. Finally, we set F2I to 24 for the final model of Fast-ACVNet and use this setting in all the following experiments.
 
 The three main differences of Fast-ACV from ACV are: 1) using lightweight backbones and aggregation network, 2) VAP and 3) F2I. In this section, we also exam the impact of each of the three components in Tab. \ref{tab:acv_to_fastacv}. To design a real-time version (i.e., runtime\textless 50ms), we follow most efficiency-oriented methods\cite{bangunharcana2021correlate, xu2021bilateral, deeppruner2019} to leverage a lightweight backbone and aggregation network. As shown in Tab. \ref{tab:acv_to_fastacv}, using the lightweight backbone and aggregation network can reduce the runtime of ACVNet from 200ms to 62ms. Integrating VAP into ACV can further reduce the runtime by 19.4\% with a small accuracy degradation. Using both VAP and F2I (24) which samples the top 24 seed candidates, i.e., Fast-ACVNet, can further accelerate the speed by $1.28\times$ with a small accuracy degradation.

\subsection{Performance of ACV and Fast-ACV} \label{sec:perform_acv_fastacv}
\noindent\textbf{ACV} An ideal cost volume should require few parameters for subsequent aggregation network and meanwhile enable a satisfactory disparity prediction accuracy. We analyze the complexity of ACV in terms of the number of parameters demanded in the subsequent aggregation network and the corresponding accuracy. We use GwcNet~\cite{guo2019group} as the baseline. In original GwcNet, it uses three stacked hourglass networks for cost aggregation. We first replace the combined volume in the original GwcNet with our ACV with other parts remaining the same. The corresponding model is denoted as Gwc-acv-3 in Tab. \ref{tab:acv_performance}. The results show that compared with GwcNet, Gwc-acv-3 improves D1 and EPE by 42.8\% and 39.5\% respectively. We further reduce the number of hourglass networks of the aggregation network from 3 to 2, 1, and 0, the correspondingly derived models are denoted as Gwc-acv-2, Gwc-acv-1 and Gwc-acv-0. The results in Tab. \ref{tab:acv_performance} show that, as the number of parameters reduced in the aggregation network, the prediction errors slightly increase. To achieve both high accuracy and efficiency, we choose Gwc-acv-2 as our final model, and we denote it as ACVNet.

\begin{table} 
\begin{center}
\caption{Comparison of ACVNet with state-of-the-art accuracy oriented methods on Scene Flow~\cite{dispNetC2016large} and ETH3D~\cite{schops2017multi}. \textbf{Bold}: Best, \underline{Underscore}: Second best.}\label{tab:acv_scene_eth}
\begin{tabular}{l|c|cc}
\hline
\multirow{2}{*}{Model} & Scene Flow & \multicolumn{2}{c}{ETH3D}\\ 
 & EPE (px) & Bad 1.0 (\%) & Bad 2.0 (\%)\\ 
\hline
PSMNet~\cite{chang2018pyramid} & 1.09& 5.02 & 1.09 \\
GANet~\cite{zhang2019ga} & 0.84 & 6.56 & 1.10 \\
CFNet~\cite{shen2021cfnet} & 0.97 & 3.31 & \underbar{0.77} \\
LEAStereo~\cite{leastereo} & 0.78 & - & - \\
HITNet~\cite{tankovich2021hitnet} & \textbf{0.43} & \underbar{2.79} & 0.80 \\
ACVNet (ours) & \underbar{0.48} & \textbf{2.58} & \textbf{0.57} \\
\hline
\end{tabular}
\end{center}
\end{table}

\begin{table} 
\begin{center}
\caption{Comparison of Fast-ACVNet with the state-of-the-art efficiency oriented methods on Scene Flow~\cite{dispNetC2016large}}\label{tab:fast_acv_scene}
\begin{tabular}{lcc}
\hline
{Model} & EPE (px) & Runtime (ms) \\ 
\hline
DeepPrunerFast~\cite{deeppruner2019} & 0.97 & 61 \\
AANet~\cite{xu2020aanet} & 0.87 & 62 \\
BGNet~\cite{xu2021bilateral} & 1.17 & 28 \\
DecNet~\cite{yao2021decomposition} & 0.84 & 50 \\
CoEx~\cite{bangunharcana2021correlate} & 0.69 & 33 \\
Fast-ACVNet (ours) & \underbar{0.64} & 39 \\
Fast-ACVNet+ (ours) & \textbf{0.59} & 45 \\
\hline
\end{tabular}
\end{center}
\vspace{-10pt}
\end{table}

\begin{table*}
    \centering
    \caption{Quantitative evaluation on the test sets of KITTI 2012~\cite{kitti2012} and KITTI 2015~\cite{kitti2015}. We split the state-of-the-art methods into two parts according to the running time whether exceeds 100ms. The results of reference methods are obtained from the official declaration. $*$ denotes the runtime is tested on our hardware (RTX 3090).}
    \label{tab:evaluation_kitti}
    \begin{tabular}{c|c|cccccc|ccc|c}
    \hline
     \multirow{3}{*}{Target} & \multirow{3}{*}{Method} & \multicolumn{6}{c|}{KITTI 2012 \cite{kitti2012}} & \multicolumn{3}{c|}{ KITTI 2015 \cite{kitti2015}}  & \\
     \cline{3-11}
    & & 3-noc & 3-all & 4-noc & 4-all & \thead{EPE \\ noc} & \thead{EPE\\all} & D1-bg & D1-fg & D1-all & \makecell{Runtime \\ (ms)} \\
    \hline
    
    \multirow{8}{*}{ \rotatebox{90}{\textit{Accuracy}}} & {GCNet} \cite{kendall2017end} & 1.77 & 2.30 &1.36 & 1.77 & 0.6 & 0.7 & 2.21 & 6.16 & 2.87 & 900 \\
    & {PSMNet \cite{chang2018pyramid}}  &1.49  & 1.89 & 1.12 & 1.42  & 0.5 & 0.6 &1.86  &4.62  &2.32  & $310^{*}$\\
    & {GwcNet~\cite{guo2019group}}  & 1.32 & 1.70 & 0.99 & 1.27 & 0.5  & 0.5 & 1.74 & 3.93 & 2.11  & $180^{*}$\\ 
    & {GANet-deep~\cite{zhang2019ga}}  & 1.19 & 1.60 & 0.91 & 1.23 & 0.4  & 0.5 & 1.48 & 3.46 & 1.81  & 180\\ 
    & CFNet~\cite{shen2021cfnet} & 1.23 & 1.58 & 0.92 & 1.18 & 0.4 & 0.5  &1.54 &3.56 &1.88 &180 \\
    & RAFT-Stereo~\cite{lipson2021raft} &- &- &- &- &- &- & 1.75 & 2.89 & 1.96 & 380 \\
    & LEAStereo~\cite{leastereo} & \textbf{1.13} & \textbf{1.45} & \textbf{0.83} & \textbf{1.08} & 0.5 & 0.5  &\underbar{1.40} &\textbf{2.91} &\textbf{1.65} &300\\
    & ACVNet (ours) & \textbf{1.13} & \underbar{1.47} & \underbar{0.86} & \underbar{1.12} & 0.4 & 0.5  &\textbf{1.37} &\underbar{3.07} &\textbf{1.65} &200\\
    \hline
    \multirow{10}{*}{ \rotatebox{90}{\textit{Speed}}}& {DispNetC \cite{dispNetC2016large}} & 4.11 & 4.65 & 2.77 & 3.20 & 0.9 & 1.0 & 4.32 & 4.41 & 4.34 & 60\\
     & {DeepPrunerFast\cite{deeppruner2019}} & - & - & - & - & - & - & 2.32 & 3.91 & 2.59 & $50^{*}$\\
     & {AANet\cite{xu2020aanet}} & 1.91 & 2.42  & 1.46 &1.87 & 0.5 & 0.6 &  1.99 & 5.39 & 2.55 & 62\\
    & {DecNet~\cite{yao2021decomposition}}  & - & - & - & - & - & - & 2.07 & 3.87 & 2.37 & 50\\
    &  {BGNet~\cite{xu2021bilateral}} & 1.77 & 2.15 &  - & - & 0.6 & 0.6 & 2.07 & 4.74  & 2.51 & $28^{*}$\\
    &  {BGNet+~\cite{xu2021bilateral}} & 1.62 & 2.03 &  1.16 & 1.48 & 0.5 & 0.6 & 1.81 & 4.09  & 2.19 & $35^{*}$\\
    &  {CoEx~\cite{bangunharcana2021correlate}} & 1.55 & 1.93 &  1.15 & 1.42 & 0.5 & 0.5 & 1.79 & 3.82  & 2.13 & $33^{*}$\\
    &  {HITNet~\cite{tankovich2021hitnet}}  & \textbf{1.41} & \underbar{1.89} &  \underbar{1.14} & \underbar{1.53} & 0.4 & 0.5 & \underbar{1.74} & 3.20  & \textbf{1.98} & $54^{*}$\\
    &  {Fast-ACVNet+ (ours)} & \underbar{1.45} & \textbf{1.85} &  \textbf{1.06} & \textbf{1.36} & 0.5 & 0.5 & \textbf{1.70} & \underbar{3.53}  & \underbar{2.01} & 45\\
    &  {Fast-ACVNet (ours)} & 1.68 & 2.13 &  1.23 & 1.56 & 0.5 & 0.6 & 1.82 & 3.93  & 2.17 & 39\\
    \hline
    \end{tabular}
\end{table*}

\begin{table} 
\begin{center}
\caption{Runtime (ms) analysis of ACVNet, Fast-ACVNet and Fast-ACVNet+ on KITTI 2015. The size of input stereo images is 1242$\times$375.}\label{tab:runtime_analysis}
\begin{tabular}{lccc}
\hline
{Module}  &ACVNet &\makecell{Fast- \\ ACVNet} & \makecell{Fast-\\ACVNet+} \\
\hline
Feature Extraction & 24 & 16 & 16 \\
ACV Construction & 58 & 8 & 14 \\
Cost  Aggregation & 85 & 9 & 9 \\
Disparity  Prediction & 33 & 6 & 6 \\
\hline
\end{tabular}
\end{center}
\end{table}

\begin{table} 
\begin{center}
\caption{Generalization evaluation on KITTI and Middlebury training sets. All models are only trained
on Scene Flow and tested on training images of two real datasets.}\label{tab:generalization}
\begin{tabular}{cccc}
\hline
Model & \makecell{KITTI 2012 \\ D1-all (\%)} & \makecell{KITTI 2015 \\ D1-all (\%)} & \makecell{Middlebury \\ Bad 2.0 (\%)} \\ 
\hline
PSMNet~\cite{chang2018pyramid} & 15.1  & 16.3 & 30.25 \\
DeepPrunerFast~\cite{deeppruner2019} & 16.8  & 15.9 & 30.83\\
BGNet~\cite{xu2021bilateral} & 24.8  & 20.1 & 37.00\\
CoEx~\cite{bangunharcana2021correlate} & 13.5  & 11.6 & 25.51\\
Fast-ACVNet (ours) & \textbf{12.4}  & \textbf{10.6} & \textbf{20.13}\\
\hline
\end{tabular}
\end{center}
\vspace{-10pt}
\end{table}

\noindent\textbf{Fast-ACV} We demonstrate the effectiveness of our attention filtering and powerful expressive ability of our Fast-ACV. As shown in Tab. \ref{tab:fast_acv_performance}, our attention filter can improve EPE from 0.83 to 0.64 when there is only one hourglass network for aggregation. The Fast-ACV without attention filter, which is a cascade cost volume, requires more hourglass aggregation networks to achieve performance gains due to the limited representation ability of the cost volume. While our Fast-ACV with only one hourglass aggregation network can achieve better accuracy than Fast-ACV without the attention filter and with three hourglass aggregation networks (i.e., the cascaded cost volume). Cascade cost volume methods reduce the memory and computational complexity of cost volume, however, for every stage, the cost volume needs to be re-constructed and re-aggregated with abundant 3D convolutions. In contrast, our Fast-ACV has a higher representation ability and thus requires only much fewer aggregation networks.

\subsection{Universality and Superiority of Our Methods} \label{sec:universality}
To demonstrate the universality and the superiority of our method, we integrate our cost volume into two state-of-the-art models, i.e. PSMNet~\cite{chang2018pyramid} and GwcNet~\cite{guo2019group}, and compare the performance of the original models with those after using our method. We denote the model after applying our method as PSM+ACV, PSM+Fast-ACV, Gwc+ACV, and Gwc+Fast-ACV respectively, as shown in Tab. \ref{tab:vs_cascade}. We also experimentally compare our ACV and Fast-ACV with the cascaded approach. We apply the two-stage cascaded method proposed by~\cite{gu2020cascade} to PSMNet and GwcNet, the corresponding model is denoted as PSM+CAS and Gwc+CAS. The results in Tab. \ref{tab:vs_cascade} show that, both our ACV and Fast-ACV can obviously improve the accuracy compared with the original model and meanwhile outperform the cascaded cost volume~\cite{gu2020cascade}. In addition, our Fast-ACV can significantly reduce the number of FLOPs, the number of parameters, and runtime compared with the original model and ~\cite{gu2020cascade}. We think the superior performance of Fast-ACV to the cascaded cost volume in terms of both accuracy and runtime is because that the latter needs to re-aggregate a cost volume for each stage without reusing the prior information from the probability volume of the previous stage. As a result, it needs three hourglass networks for aggregation. In comparison, our Fast-ACV is more informative and expressive, which only needs an hourglass network for aggregation. 

It is noteworthy that our VAP module can also be easily integrated into stereo networks based on linear up-sampling cost volume in disparity prediction stage, such as PSMNet and GwcNet. As shown in Tab. \ref{tab:vap}, we integrate our VAP into PSMNet and GwcNet to derive PSMNet-VAP and GwcNet-VAP respectively. We observe that our VAP can achieve significant performance improvements with little extra computation.

\subsection{Comparisons with State-of-the-art}
\noindent\textbf{Scene Flow and ETH3D} As shown in Tab. \ref{tab:acv_scene_eth}, our ACVNet achieves the
state-of-the-art performance. We can observe that our ACVNet improves EPE accuracy by 38.4\% on Scene Flow with the state-of-the-art method LEAStereo~\cite{leastereo}. On ETH3D, our ACVNet also achieves state-of-the-art results. On Scene Flow, our Fast-ACVNet+ achieves the remarkable EPE of 0.59, which ourperforms all other real-time methods~\cite{deeppruner2019,xu2020aanet,xu2021bilateral,yao2021decomposition,bangunharcana2021correlate} at time of writing. Representative competitors are reported in Tab. \ref{tab:fast_acv_scene}. Qualitative results are shown in Fig. \ref{fig:sceneflow}. Compared with Fast-ACVNet, Fast-ACVNet+ constructs correlation volume at 1/4 resolution.

\noindent\textbf{KITTI} As shown in Tab. \ref{tab:evaluation_kitti}, our ACVNet outperforms most existing published methods and achieves comparable accuracy with LEAStereo, but is faster than it, i.e., 200ms vs 300ms. Meanwhile, our Fast-ACVNet+ outperforms all the published real-time methods (i.e., inference time is smaller than 50ms) on the KITTI 2012 and 2015 benchmarks. Qualitative results are shown in Fig. \ref{fig:kitti}. Fast-ACVNet+ also achieves comparable accuracy with HITNet~\cite{tankovich2021hitnet}, but with a faster inference speed, i.e., 45ms vs 54ms. In order to ensure the fairness of the comparison, the runtime of HITNet is tested on our hardware (RTX 3090) using the open-source models in PyTorch.

\noindent\textbf{Runtime analysis} We also report the runtime breakdown of our ACVNet, Fast-ACVNet, and Fast-ACVNet+ based on images of KITTI 2015 whose image size is 1242 $\times$ 375. Compared with Fast-ACVNet, Fast-ACVNet+ requires an additional 6ms due to constructing a 1/4 resolution correlation volume for Fast-ACV. 

\begin{figure}
\begin{center}
\includegraphics[width=1.0\linewidth]{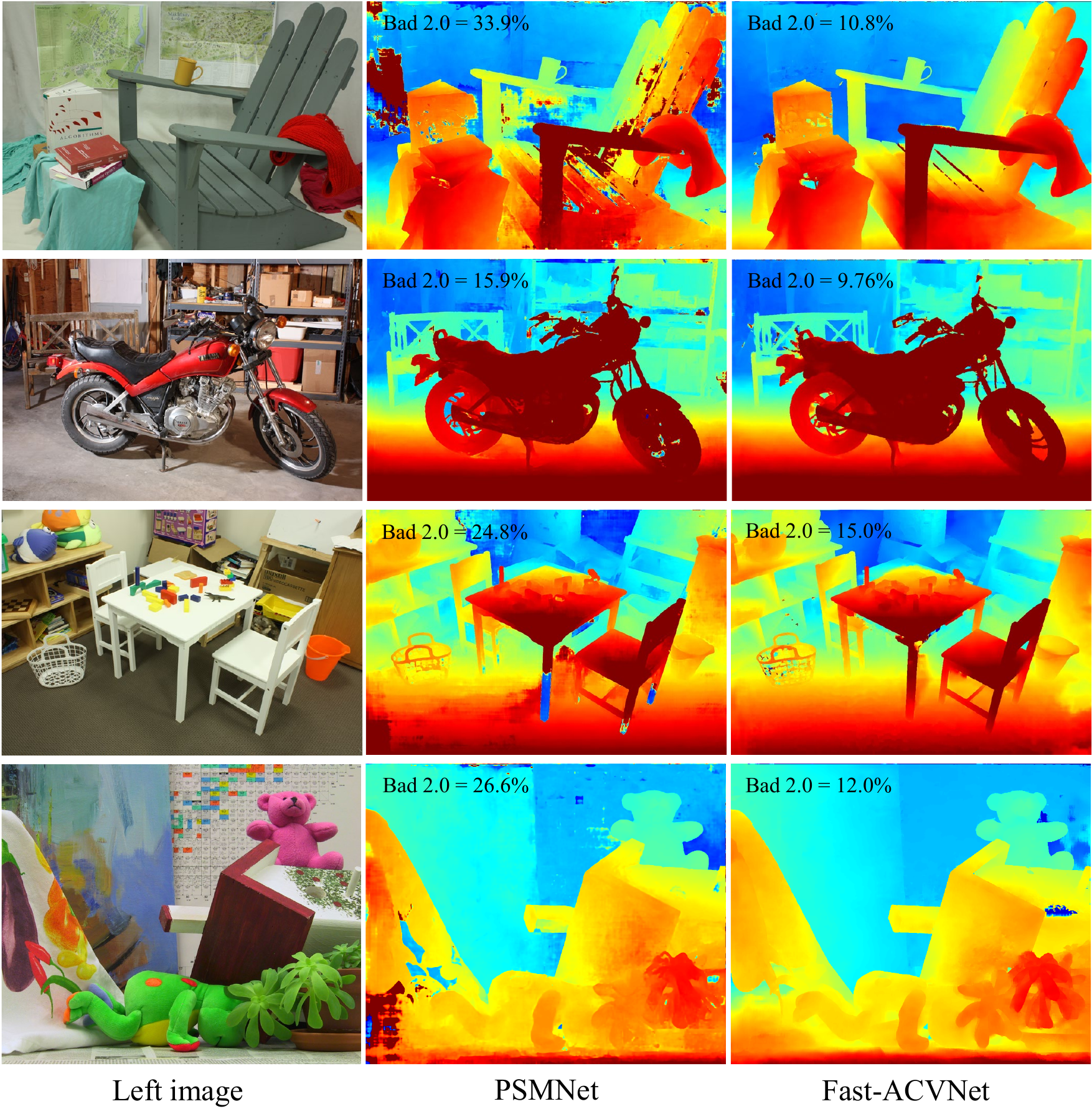}
\end{center}\caption{Qualitative results of generalization performance evaluation on the Middlebury 2014 dataset.}
\label{fig:middle}
\end{figure}

\subsection{Generalization Performance} \label{sec:generalization}
In addition to impressive performance on Scene Flow, ETH3D and KITTI, we also evaluate the generalization ability of our methods on the half-resolution training set of the Middlebury 2014 dataset~\cite{scharstein2014high} and the full-resolution training set of the KITTI 2012 and 2015. In this evaluation, all the comparison methods are only trained on the Scene Flow. Qualitative results are shown in Fig. \ref{fig:middle}. As shown in Tab. \ref{tab:generalization}, our Fast-ACVNet achieves state-of-the-art performance and outperforms other real-time methods.

\section{Conclusion}
In this paper, we propose a novel cost volume construction method, named attention concatenation volume (ACV), and its real-time version Fast-ACV, which generates attention weights based on similarity measures to filter the concatenation volume. Based on ACV and Fast-ACV, we design a highly accurate network ACVNet and real-time network Fast-ACVNet, which achieve state-of-the-art performance on several benchmarks (i.e., our ACVNet ranks the $2^{nd}$ on KITTI 2015 and Scene Flow, and the $3^{rd}$ on KITTI 2012 and ETH3D; our Fast-ACVNet outperforms almost all the state-of-the-art real-time methods on Scene Flow, KITTI 2012 and 2015 and meanwhile has the best generalization ability among all real-time methods). Our ACV and Fast-ACV are general cost volume representations that can be integrated into many existing stereo matching models based on 4D cost volumes for performance improvement. In addition, the idea of our Fast-ACV, which exploits prior information in the probability volume of the previous stage as attention weights to filter the cost volume of the current stage, can be also applied to the cascade cost volume methods to achieve accuracy gain and greatly reduce the amount of computation and parameters.

\vspace{2mm}
\noindent\textbf{Acknowledgements.} This work was supported in part by the National Natural Science Foundation of China under Grants 62122029, 62061160490 and U20B2064.

\ifCLASSOPTIONcaptionsoff
  \newpage
\fi

{\small
    \bibliographystyle{IEEEtran}
    \bibliography{Fast_ACV}
}

\begin{IEEEbiography}[{\includegraphics[width=1in,height=1in,clip,keepaspectratio]{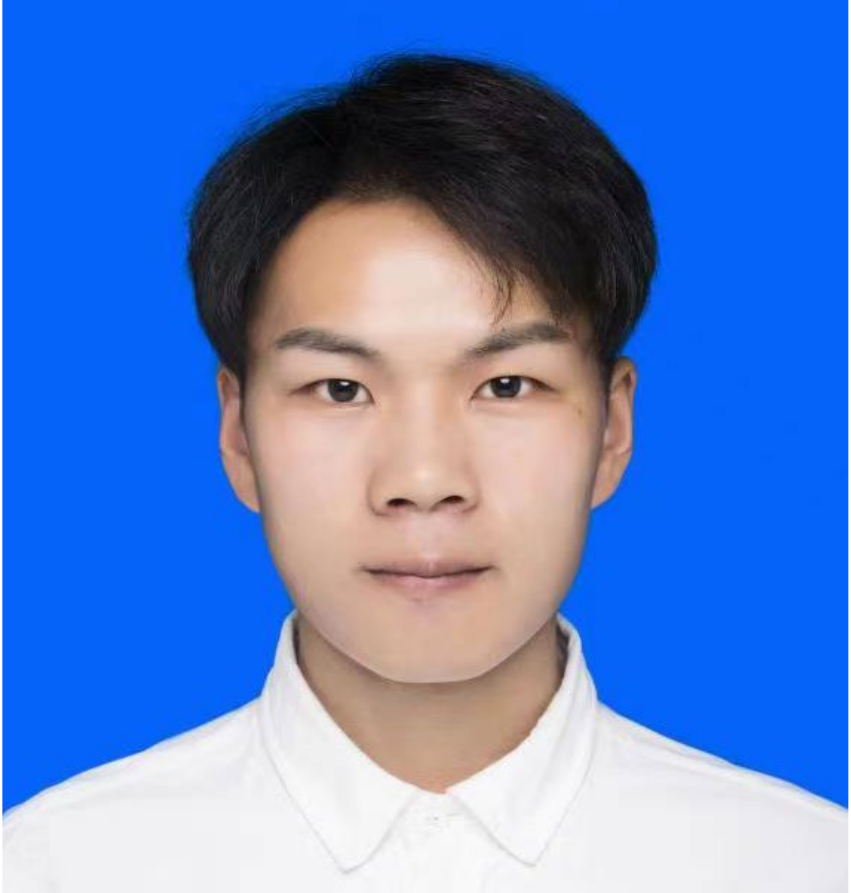}}]{Gangwei Xu}
  is a PhD student at the Department of Electronic Information and Communications at Huazhong University of Science and Technology. He is supervised by Prof. Xin Yang. He received the B.Eng. degree from Huazhong University of Science and Technology in 2021. His research interests include stereo matching and optical flow estimation. He has published two papers in CVPR 2022 and CVPR 2023.
\end{IEEEbiography}
\vspace{-10pt}



\vspace{-30pt}
\begin{IEEEbiography}[{\includegraphics[width=1in,height=1in,clip,keepaspectratio]{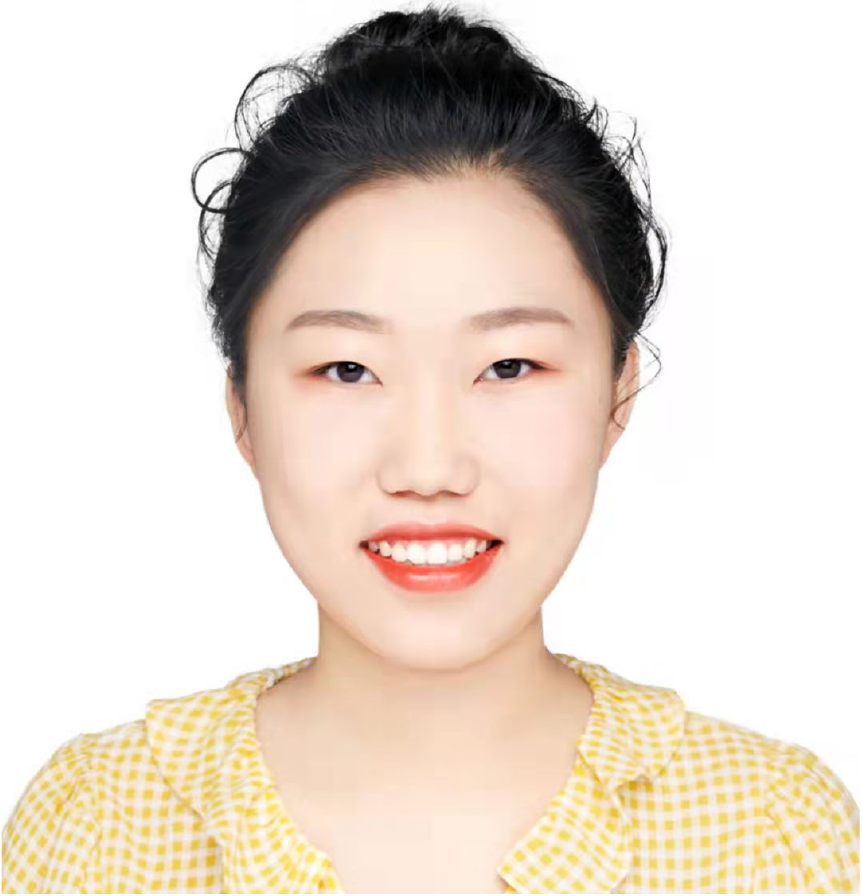}}]{Yun Wang}
  is a PhD student at the Department of Electronic Information and Communications at Huazhong University of Science and Technology. She is supervised by Prof. Xin Yang. She received the B.Eng. degree from Huazhong University of Science and Technology in 2020. Her research interests include stereo matching, optical flow estimation and scene flow estimation.
\end{IEEEbiography}
\vspace{-10pt}

\vspace{-10pt}
\begin{IEEEbiography}[{\includegraphics[width=1in,height=1in,clip,keepaspectratio]{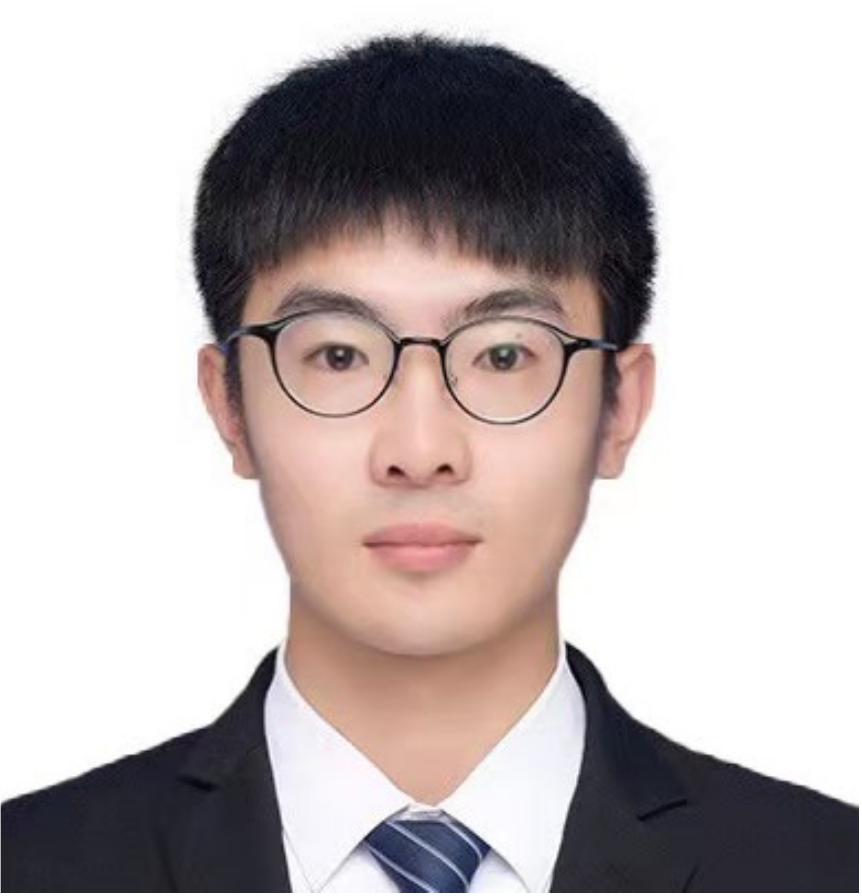}}]{Junda Cheng}
  is a PhD student at the Department of Electronic Information and Communications at Huazhong University of Science and Technology. He is supervised by Prof. Xin Yang. He received the B.Eng. degree from Huazhong University of Science and Technology in 2020. His research interests include stereo matching and 3D vision.
\end{IEEEbiography}
\vspace{-10pt}

\vspace{-10pt}
\begin{IEEEbiography}[{\includegraphics[width=1in,height=1in,clip,keepaspectratio]{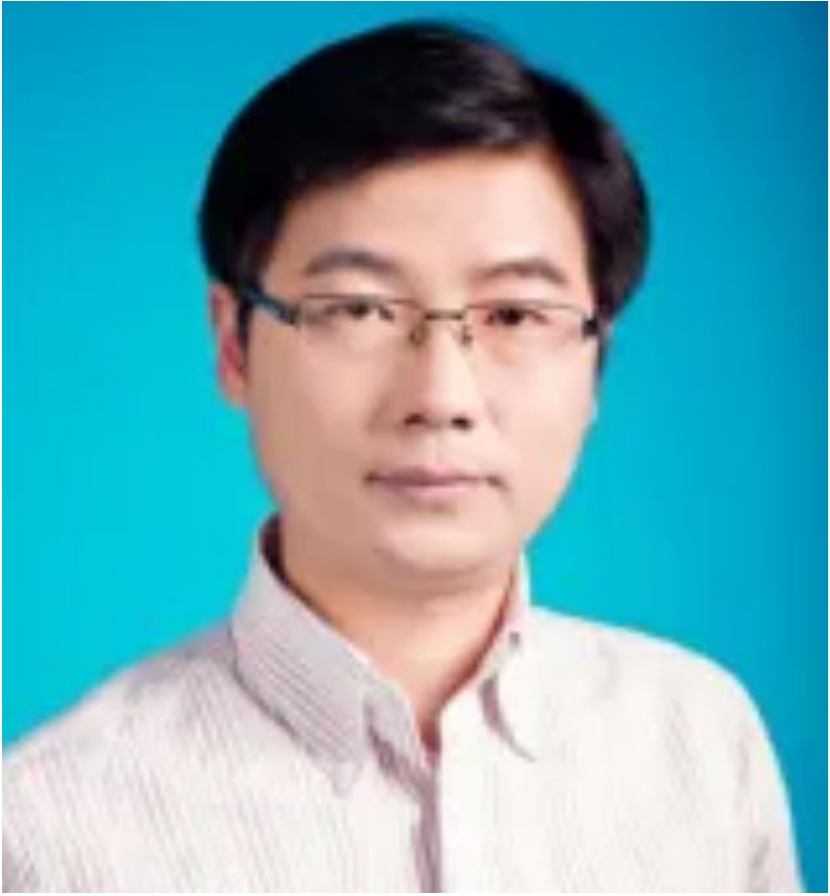}}]{Jinhui Tang}
  is a Professor at the Nanjing University of Science and Technology. He received the B.Eng. and Ph.D. degrees from the University of Science and Technology of China in 2003 and 2008, respectively. He has authored over 150 papers in top-tier journals and conferences, with more than 10,000 citations in Google Scholar. His research interests include multimedia analysis and computer vision. He was a recipient of the best paper awards in ACM MM 2007, PCM 2011 and ICIMCS 2011, the Best Paper Runner-up in ACM MM 2015, and the best student paper awards in MMM 2016 and ICIMCS 2017. He has served as an Associate Editor for the IEEE TNNLS, IEEE TKDE, IEEE TMM and IEEE TCSVT.
\end{IEEEbiography}
\vspace{-10pt}

\vspace{-30pt}
\begin{IEEEbiography}[{\includegraphics[width=1in,height=1in,clip,keepaspectratio]{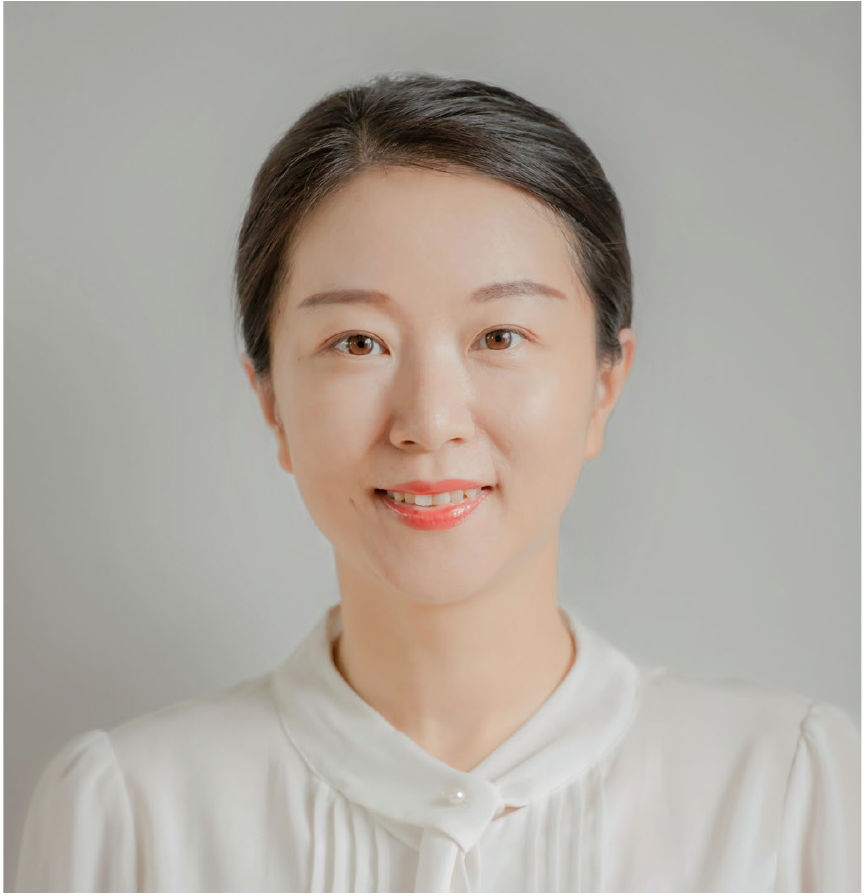}}]{Xin Yang}
   is a Professor at the Department of Electronic Information and Communications at Huazhong University of Science and Technology. She received her Ph.D. degree in the Department of Electrical Computer Engineering at the University of California, Santa Barbara (UCSB).  Her research interests include medical image analysis and 3D vision. She is the recipient of the National Natural Science Fund of China for Excellent Youth Scholar and China Society of Image and Graphics Qingyun Shi Female Scientist Award. She has published over 90 technical papers and held 12 patents. She serves as an Associate Editor of IEEE-TMI and Multimedia System, an Area Chair of MICCAI’19-21 and ACM MM’18, and a PC member of CVPR, ECCV and ICCV. She is also a reviewer of top journals such as IEEE-TPAMI, IEEE-TNNLS, MedIA, etc.
\end{IEEEbiography}
\vspace{-10pt}








\end{document}